\documentclass{article}
\usepackage{geometry}
\usepackage{authblk}

\usepackage{hyperref}
\usepackage{tikz}
\usepackage{color}
\usepackage{amsmath,amsthm,amssymb}
\usepackage{xspace}
\usepackage{comment}

\usepackage{arydshln}
\newcommand\newcolor[2]%
  {\definecolor{#1}{RGB}{#2}%
   \expandafter\newcommand\csname#1\endcsname[1]{\textcolor{#1}{##1}}%
  }
  
\newcommand{\abm}{ABEM\xspace}
\newcommand{\waningmodel}{IWM\xspace}
\newcommand{\agemodel}{ASM\xspace}
\newcommand{\hospitalmodel}{HM\xspace}

\newcommand{\abmLong}{Agent-Based Epidemics Model\xspace}
\newcommand{\waningmodelLong}{Immunity Waning Model\xspace}
\newcommand{\agemodelLong}{Age Structure Model\xspace}
\newcommand{\hospitalmodelLong}{Hospitalisation Model\xspace}

\newcommand{\sst}[1]{Success Story #1\xspace}

\newcommand{\cldcaption}[1]{Schematic causal loop diagram of all elements regarded in the four models discussed in this work. The greyed out components are not regarded in the #1. The components coloured in black represent the model states, the components coloured in green the model inputs and the components coloured in blue the model outputs.\xspace}

\usetikzlibrary{shapes,decorations,arrows,calc,arrows.meta,fit,positioning,backgrounds}

\AtBeginDocument{%
  \providecommand\BibTeX{{%
    \normalfont B\kern-0.5em{\scshape i\kern-0.25em b}\kern-0.8em\TeX}}}

\title{Model Families for Multi-Criteria Decision Support: A COVID-19 Case Study}

\author[1,2]{Martin Bicher}
\author[2]{Claire Rippinger}
\author[1,2]{Christoph Urach}
\author[1,2]{Dominik Brunmeir}
\author[2]{Melanie Zechmeister}
\author[1,2,3]{Niki Popper}

\affil[1] {TU Wien, Institute of Information Systems Engineering, Favoritenstra\ss{}e 9-11, 1040, Vienna, Austria}
\affil[2]{dwh GmbH,Neustiftgasse 57-59, 1070 Vienna,  Austria}
\affil[3]{Association for Decision Support in Health Policy and Planning, Neustiftgasse 57-59, 1070 Vienna,  Austria}
\begin{document}
\maketitle
\begin{abstract}
 Continued model-based decision support is associated with particular challenges, especially in long-term projects. Due to the regularly changing questions and the often changing understanding of the underlying system, the models used must be regularly re-evaluated, -modelled and -implemented with respect to changing modelling purpose, system boundaries and mapped causalities. Usually, this leads to models with continuously growing complexity and volume. In this work we aim to reevaluate the idea of the model family, dating back to the 1990s, and use it to promote this as a mindset in the creation of decision support frameworks in large research projects. The idea is to generally not develop and enhance a single standalone model, but to divide the research tasks into interacting smaller models which specifically correspond to the research question. This strategy comes with many advantages, which we explain using the example of a family of models for decision support in the COVID-19 crisis and corresponding success stories. We describe the individual models, explain their role within the family, and how they are used -- individually and with each other.
\end{abstract}

\paragraph{Keywords.}model family, decision support framework, covid-19, sars-cov-2

\maketitle

\section{Introduction}
Developing the right computer model for a specific purpose is crucial for good modelling practices, regardless of the area of application. This principle can be found in various modelling guidelines and tutorials \cite{crout2008chapter,roberts2012conceptualizing}. This pragmatism (compare Stachowiak, definition "model" \cite{stachowiak1973allgemeine}) refers only secondarily to the nature of the abstracted system but primarily to the questions to be answered about the system. It involves selecting the modelling method, database, in- and output, system variables, and resolution of the model.
Especially, modelling in long-lasting decision support projects is challenging due to the constant need to modify the decision framework based on new tasks and information about the system.

The most straight forward solution to this problem is to \textbf{extend} or modify the one existing decision support model. This strategy is usually the quickest, but also the riskiest:  If one retains or extends the model for too long one ``may extrapolate beyond the region of fit'' or ``draw 33rd-order conclusions from a 1st-order model'', both rendering the model \textit{invalid} for the given purpose (we used the terminology of Golomb's famous ``Do's and Don'ts of Mathematical Modelling'' \cite{golomb1971mathematical}). Moreover, if an existing model was extended beyond a certain complexity, it becomes \textit{inflexible} due to long computation times and high number of model parameters. This causes problems related to \textit{sensitivity, verification} and \textit{validation}. Finally, also model \textit{documentation} and thus model \textit{communication} becomes continuously more difficult .

To solve the problem, one can completely redevelop and \textbf{replace} the model for the new use case. This is costly but avoids problems with existing limitations. However, it requires developing, validating, and verifying a new model, as well as ensuring compatibility with the old one for the sake of  \textit{validity}, \textit{credibility} and \textit{reproducibility} of the old results. 

In this work, we advertise a different mindset for model development: instead of replacing an old model with a new one, the new model can be seen as an addition to an entire pool of models, henceforth referred to as a \textbf{model family}, a term which was, to the authors' knowledge, first introduced by P.K. Davis in the 1990s in a slightly different context\cite{MR-1004-DARPA,R-4252-DARPA}. Hereby we refer to a collection of different interacting models with different fields of applications, model boundaries and resolutions. Instead of attempting to answer every decision-relevant question using the same model, the questions are distributed to the most suitable model(s) in the family.


Between 2020 and 2023 a team of dwh GmbH and TU Wien provided decision support for Austrian policy makers and health care institutions on the subject of the SARS-COV-2 crisis. The team faced many challenges during these years, including a quickly growing knowledge base, a continuously evolving system and the constantly changing needs of the decision makers. To keep up with these changes, the team developed a family of seven different models in total. The four most relevant will be discussed in detail in this work.


We will show the development and usage of the four decision support models in Austria and present success stories that illustrate the benefits of a model family versus a standalone model. Additionally, we will discuss the use of Causal Loop Diagrams (CLD) to visualise and analyse the relationships and roles of the models within the family.

Aim of this work is to demonstrate the advantages and challenges of creating and utilising model families. We want to encourage modellers working on complex issues like COVID-19 to prioritise developing a family of models instead of constantly improving a standalone model.

\section*{Methods}
This chapter will describe our interpretation of the term \textit{model family} and how we used the concept of Causal-Loop-diagrams to visualise the role of a model within its family. Moreover we will present the motivation and development of our COVID-19 model family, provide an introduction to each model, and an overview of their specifications, parameters, and implementation. Details can be found in the Appendix or previously published material.

\subsection*{General Model Family Concept}
Based on the work in \cite{MR-1004-DARPA,R-4252-DARPA}, we define a \textit{model family} as a collection of different models with model different aspects of one large overall system. The models may have different
\begin{itemize}
\item modelling approaches,
\item resolutions,
\item modelling purposes,
\item model boundaries,
\item regions of validity, and
\item time-frames of validity. 
\end{itemize}
The last refers to the problem that changing knowledge base of the overall system might render an existing model at least partially less valid due to novel information. By the term resolution we refer to temporal scale, spatial scale, process detail, object-related structure and system structure, as specified in \cite{MR-1004-DARPA}. 

Anyway, a family is well designed, if (a) any subsystem in the regarded overall system is covered by at least one model and (b) any two models differ by at least one of the aforementioned points. It is typically seen in context with its genesis and further development: Enhancement can take place by extending existing models, adding completely new ones which cover areas and questions previously not included, and also by dividing existing models into individual submodels to enhance their flexibility. 

We want to emphasise, that a \textit{model family}, in our understanding, does not rely on automated coupling of the models, neither interfaced, integrated nor sequential (see \cite{SWINERD2012118}). This clearly distinguishes the concept from \textit{multi-method modelling} \cite{balaban2015toward},  \textit{co-simulation} \cite{hafner2017terminology} or \textit{multi resolution modelling} \cite{MR-1004-DARPA}, or \textit{hybrid simulation} (different definitions, e.g. \cite{BRAILSFORD2019721}, \cite{pian1973hybrid}). Moreover, the individual models in the family may follow different modelling purposes and goals. Therefore, they are not only versions of the same model with different resolution, which poses a difference to Davis' ideas of a \textit{variable resolution-}, \textit{multi resolution-} or \textit{cross resolution model family}\cite{R-4252-DARPA}.

\subsection*{Mapping Models with Causal Loop Diagrams}
The causal loop diagram concept was developed together with the modelling approach System Dynamics by J.W. Forrester in the 1970s \cite{forrester1970urban,forrester1987lessons}. The concept originally analyses causal relations and loops within a system to develop a System Dynamics model. The diagram uses nodes and directed edges to represent components and their relationships. Edges are labelled with signs indicating whether the causality acts reinforcing or balancing.

In our work, we put these diagrams in an entirely different context (e.g. Figure \ref{fig:cld_abm}). Instead of using the diagram of the observed system to generate a model, we instead mark the system components and causal relations and loops covered by existing models, i.e. the models in our model family. For this use, we defined the following convention: 
\begin{itemize}
    \item Nodes representing system components which are not depicted in the model are coloured light-grey.
    \item Nodes representing modelled system components distinguished with respect to their role in the model: inputs are coloured green, state variables black, and outputs blue.
    \item Edges representing modelled causal relations are drawn in black, others are coloured light-grey.
\end{itemize}

\subsection*{COVID-19 Model Family Development}
The COVID crisis is a perfect example for a highly complex, continuously evolving system with changing knowledge base and changing needs for decision support:

In early 2020, decision makers were primarily interested in scenario forecasts for the potential impact of the new virus on the population and the health care system. Due to the quick spread of the disease in spring 2020, the need for non pharmaceutical intervention modelling arose for policy making.
Models had to be quickly adapted to new research on virus parameters, treatments, and vaccines. As immune escape variants emerged and vaccinations became widely available, models had to be extended to include population immunity. Finally, in mid-2022, decision makers required long-term analysis of the system to evaluate exit strategies.
These changes led to new modelling challenges, requiring changes to modelling purpose, boundaries, and causal relations. Table \ref{tab:my_label} in the Appendix provides a timeline of the changes and the team's modifications to their model family. We clearly see, that the \abm played the most important role in the process, yet was not extended beyond a certain region of validity and usability. Instead, other models were added to supplement. 

The first model in the family is a large-scale epidemiological agent-based model (\abmLong, short \abm). It was also the model of the family that was first developed and covers most components of the overall system. Therefore it is often taken as a reference. The second model (\waningmodelLong, short \waningmodel), deals with the immunity of the population, the third model (\hospitalmodelLong, short \hospitalmodel) depicts hospitalisations, and the fourth model (\agemodelLong, short \agemodel) solely regards the age-distribution of cases during an epidemic wave.


\subsection*{\abmLong}
\label{sec:agent_based_overall_model}
\begin{figure}
\centering
\begin{tikzpicture}[-Latex,auto,node distance =1 cm and 1 cm,semithick,
state/.style args={#1}{rounded rectangle, draw, minimum width = 2 cm,fill=#1!10,text=#1,draw=#1, align = center},
edgecolor/.style args={#1}{color=#1},
point/.style = {circle, draw, inner sep=0.04cm,fill,node contents={}},
bidirected/.style={Latex-Latex,dashed},
el/.style = {inner sep=2pt, align = left, sloped},
every edge/.append style={nodes={pos=0.25,anchor=center, circle, draw,fill=white,font = \tiny, inner sep=0.04cm}}]
\input{clds/colordefs}
\def \diC {outputColor};
\def \uiC {outputColor};
\def \testsC {inputColor};
\def \varC {inputColor};
\def \seasC {inputColor};
\def \contC {varColor};
\def \polC {inputColor};
\def \infC {varColor};
\def \infedC {outputColor};
\def \recC {varColor};
\def \immC {outputColor};
\def \susC {varColor};
\def \vaccC {inputColor};
\def \lostC {varColor};
\def \hospC {ignoreColor};
\def \deadC {varColor};
\def \icuC {ignoreColor};
\def \nonicuC {ignoreColor};
\def \transicuC {ignoreColor};
\def \transnonC {ignoreColor};
\def \vartwoC {inputColor};
\def \varthreeC {inputColor};

\def \uiinfedC {linkColor};
\def \diinfedC {linkColor};
\def \transuiC {linkColor};
\def \transdiC {linkColor};
\def \seasuiC {linkColor};
\def \seasdiC {linkColor};
\def \testsdiC {linkColor};
\def \testsuiC {linkColor};
\def \contuiC {linkColor};
\def \contdiC {linkColor};
\def \polcontC {linkColor};
\def \suscontC {linkColor};
\def \immcontC {linkColor};
\def \infcontC {linkColor};
\def \vaccsusC {linkColor};
\def \vaccimmC {linkColor};
\def \lostsusC {linkColor};
\def \immlostC {linkColor};
\def \immsusC {linkColor};
\def \recimmC {linkColor};
\def \infedrecC {linkColor};
\def \infedinfC {linkColor};
\def \infedhospC {ignoreColor};
\def \infeddeadC {linkColor};
\def \hospdeadC {ignoreColor};
\def \hospicuC {ignoreColor};
\def \hospnonicuC {ignoreColor};
\def \nonicutransicuC {ignoreColor};
\def \transnonnonicuC {ignoreColor};
\def \icutransnonC {ignoreColor};
\def \transicuicuC {ignoreColor};
\def \transicunonicuC {ignoreColor};
\def \transnonicuC {ignoreColor};
\def \varuiC {linkColor};
\def \vardiC {linkColor};
\def \vartwolostC {linkColor};
\def \vartwoimmC {linkColor};
\def \varthreedeadC {linkColor};
\def \varthreehospC {ignoreColor};

\input{clds/cld_base}
\end{tikzpicture}
\caption{\cldcaption{\abmLong}}
\label{fig:cld_abm}
\end{figure}
The \abmLong  (henceforth \abm) was the first and most complex member of the model family to be implemented. Adapted from an existing model to simulate the spread of influenza, it uses a population of agents and a contact network to model the spread of an infectious disease. As it reproduces the demographic of Austria and explicitly models \textit{contact-locations} such as households, schools and work places, it has many fields of application:
\begin{itemize}
\item forecasting of infections (COVID Forecasting Consortium \cite{bicher2022supporting})
\item evaluation policies (tracing methods \cite{Bicher_Evaluation_2021})
\item better understanding several aspects of the pandemic (undetected cases \cite{Rippinger_Evaluation_2021}, immunity waning, and their impact on the herd immunity \cite{bicher2022model})
\item support for other logistical and strategic decisions (test logistics \cite{Wolfinger_Large_2021}, vaccination program \cite{Jahn_Targeted_2021,Bicher_Iterative_2022}, wastewater surveillance of virus variants \cite{Amman_Viral_2022})
\item source for synthetic epidemic data \cite{Popper_Synthetic_2021}
\end{itemize}
The model was also a cornerstone of many other commissioned modelling studies which were not published in peer reviewed journals (see \href{https://www.dwh.at/en/projects/covid-19/}{https://www.dwh.at/en/projects/covid-19/} for details).

Figure \ref{fig:cld_abm} shows a schematic causal loop diagram of all elements and interactions regarded by the four models discussed in this work.


\subsubsection*{Short Model Description}
The \abm is an agent-based SEIR-model (susceptible--exposed--infectious--recovered, see \cite{brauer2008compartmental}). Every inhabitant of the country is depicted as an agent with certain sex, age and residence place (coordinate). According to regional and socio-demographic structure, agents are assigned contact locations (households, school classes, workplaces, care homes) where they are able to meet other agents. In case of a contact between susceptible and infectious agents, an infection occurs with a probability depending on many epidemiological factors, such as virus strain, seasonality, location, shedding, and adherence. Infected agents then follow a disease progression path including relevant events from infection to immunity loss: start of infectiousness, symptom onset, recovery or death, start of immunity to immunity loss. While interactions between agents are evaluated in discrete time steps of one day, the disease progression is simulated using a discrete event strategy.

The most comprehensive parts of the model are related to the implementation of policies including symptomatic/screening tests, quarantine, contact tracing, vaccinations, school/workplace closure, and increased awareness. Imported cases (tourism) and introduction of new variants are handled by random external infections.

Its original full model specification including parameter values was published in \cite{Bicher_Evaluation_2021}. Since the model and its parametrisation is constantly updated to the newest information, its most recent version can be found on the homepage of dwh GmbH (\href{https://www.dwh.at/en/projects/covid-19/}{https://www.dwh.at/en/projects/covid-19/}, section ``Technical documents, further information and resources'').

\subsubsection*{Model Usage}


For most usages, the model is calibrated to match the historical number of reported infections. This way, a population of agents is produced which matches the current Austrian population with regard to active or past infections and immunity. The model can than be computed into the future to make short or medium term forecasts or to analyse several scenarios which simulate varying strategies (e.g. different test concepts, lockdown strategies, vaccination programs, ...) or uncertain systemic events (introduction of new variants, immunity against new variants, ...).

Additionally, the simulated population can be used to analyse the past course of the pandemic with regard to the proportion of undetected cases or infection networks.

\subsubsection*{Parametrisation and Calibration}
The model utilises an enormous number of over 30 different partially time-, partially location- dependent model parameters. Values for these were taken from literature, surveillance data, census data or were guessed by domain experts. Population data is mostly taken from the Austrian Bureau of Statistics, same holds true for data for contact locations. Contacts themselves are parametrised using data from the POLYMOD survey \cite{mossong_polymod_2017} and mobile phone data (origin-destination matrices). Disease and immunisation data is collected from literature and the national epidemiological surveillance system. Vaccination data is taken from aggregated exports of the Austrian electronic health record.

The ground truth for the calibration are reported confirmed cases which are matched with the outcomes of the symptomatic and screening tests in the model. The free variables of the calibration process are (mainly) parameters related to the efficiency of policies.

\subsubsection*{Implementation and Source Code}
The model is implemented in JAVA based on our own agent-based simulation tool (Agent-Based Template, ABT, \cite{abt_hp}). One simulation run for Austria with roughly 9Mio agents requires about 30-40GB RAM and 15-30sec per simulation-day (i.e. 1.5-3h per simulation-year).
Since the source code (a) is huge with more than 150 Java classes, (b) subject to constant updates, (c) partially uses parametrisation data subject to privacy, and (d) cannot be cleaned and prepared for the scientific community with feasible effort, it is not open access. It can be shared in scientific collaborations though.

\subsection*{\waningmodelLong}
\label{sec:immunity_waning_model}
\begin{figure}
\centering
\begin{tikzpicture}[-Latex,auto,node distance =1 cm and 1 cm,semithick,
state/.style args={#1}{rounded rectangle, draw, minimum width = 2 cm,fill=#1!10,text=#1,draw=#1, align = center},
edgecolor/.style args={#1}{color=#1},
point/.style = {circle, draw, inner sep=0.04cm,fill,node contents={}},
bidirected/.style={Latex-Latex,dashed},
el/.style = {inner sep=2pt, align = left, sloped},
every edge/.append style={nodes={pos=0.25,anchor=center, circle, draw,fill=white,font = \tiny, inner sep=0.04cm}}]
\input{clds/colordefs}
\def \diC {inputColor};
\def \uiC {varColor};
\def \testsC {ignoreColor};
\def \varC {ignoreColor};
\def \seasC {ignoreColor};
\def \contC {varColor};
\def \polC {ignoreColor};
\def \infC {ignoreColor};
\def \infedC {varColor};
\def \recC {varColor};
\def \immC {outputColor};
\def \susC {outputColor};
\def \vaccC {inputColor};
\def \lostC {varColor};
\def \hospC {ignoreColor};
\def \deadC {ignoreColor};
\def \icuC {ignoreColor};
\def \nonicuC {ignoreColor};
\def \transicuC {ignoreColor};
\def \transnonC {ignoreColor};
\def \vartwoC {inputColor};
\def \varthreeC {ignoreColor};

\def \uiinfedC {linkColor};
\def \diinfedC {linkColor};
\def \transuiC {linkColor};
\def \transdiC {linkColor};
\def \seasuiC {ignoreColor};
\def \seasdiC {ignoreColor};
\def \testsdiC {ignoreColor};
\def \testsuiC {ignoreColor};
\def \contuiC {linkColor};
\def \contdiC {ignoreColor};
\def \polcontC {ignoreColor};
\def \suscontC {ignoreColor};
\def \immcontC {ignoreColor};
\def \infcontC {ignoreColor};
\def \vaccsusC {linkColor};
\def \vaccimmC {linkColor};
\def \lostsusC {linkColor};
\def \immlostC {linkColor};
\def \immsusC {linkColor};
\def \recimmC {linkColor};
\def \infedrecC {linkColor};
\def \infedinfC {ignoreColor};
\def \infedhospC {ignoreColor};
\def \infeddeadC {ignoreColor};
\def \hospdeadC {ignoreColor};
\def \hospicuC {ignoreColor};
\def \hospnonicuC {ignoreColor};
\def \nonicutransicuC {ignoreColor};
\def \transnonnonicuC {ignoreColor};
\def \icutransnonC {ignoreColor};
\def \transicuicuC {ignoreColor};
\def \transicunonicuC {ignoreColor};
\def \transnonicuC {ignoreColor};
\def \varuiC {ignoreColor};
\def \vardiC {ignoreColor};
\def \vartwolostC {linkColor};
\def \vartwoimmC {linkColor};
\def \varthreedeadC {ignoreColor};
\def \varthreehospC {ignoreColor};

\input{clds/cld_base}
\path (di) edge[bend left=15,edgecolor=linkColor,dashed] node {$+$}  (cont);
\end{tikzpicture}
\caption{\cldcaption{\waningmodelLong} The dashed arrow indicates a causal link which is implemented inversely in the model.}
\label{fig:cld_waningmodel}
\end{figure}
Due to the long course of the pandemic and the emergence of new virus variants, research on immunity and in particular immunity waning became more and more relevant. Since the number of immunised persons has massive implications on the progress on epidemic waves, estimates for this quantity became an important variable of interest. Although the \abm is fully capable of giving estimates for this number (e.g. see \cite{rippinger2021evaluation,bicher2022model}) long computation times limit its capabilities to experiment with different waning distributions. The \waningmodelLong (\waningmodel) was developed to overcome this problem. Focusing only on the past and current situation, the model does not include classic epidemiological mechanisms like infections, but treats them as inputs. This leads to much smaller computation times and improved capabilities for parameter studies.

\subsubsection*{Short Model Description}



The \waningmodel itself is conceptualised based on the idea that the immunisation level against a certain virus variant is solely dependent on past infections and vaccinations. 

As displayed in Figure \ref{fig:cld_waningmodel}, the model uses this historical data as input and creates \textit{immunisation-events} which are then distributed among the entities. To get a correct picture of the overall immunity, the officially confirmed infection numbers are not sufficient because not all actual infections are getting detected e.g. due to a lack of symptoms. To solve this problem model applies an estimate for the detection rate (taken from literature with corresponding studies) to compute an estimate for the overall infection count from the detected infections. This is indicated by the dashed arrow in Figure \ref{fig:cld_waningmodel}. Undetected infections are furthermore treated and distributed analogously to the detected ones. The distribution process is deliberately kept very simple: the events are distributed randomly among the subset of eligible entities, regardless of age, gender, or other personal properties. An entity is considered eligible for an infection-based immunisation-event if they are not already labelled as \textit{immune} and they may be assigned a first/second/third/… vaccination-based immunisation-event if they have already received no/one/two/… shots with sufficient time between the shots.

Once an entity has been assigned an immunisation-event, they gain immunity – in specific immunity against infection by an observed SARS-COV-2 variant – with a given probability. In case the entity has been labelled as immune, an \textit{immunity-loss} event is scheduled after a certain amount of time drawn randomly from a previously defined distribution.

In order to evaluate ``immunity'' against a e.g. severe disease progression (hospitalisation) a second \textit{immune} state is introduced for which different distributions are used.

A detailed model specification is found in \ref{sec:spec_waningmodel}.

\subsubsection*{Model Usage}
The models' usage can be split into four areas. First the model can be used to estimate the current and past immunisation level against infection. This can be valuable to get an idea of the immunisation level necessary for natural peaks of disease waves. Secondly, the model can be used to estimate the future dynamics of the current immunisation level without regarding any future infections or vaccinations. Thirdly, the model can be applied to forecasts of case-numbers and/or vaccination numbers generated by other models to estimate the immunisation level during an upcoming epidemic wave (see \sst{3}). Finally, the model can also be used for communication purposes showing differences between immunity against infection and ``immunity''\footnote{In this case, this should be interpreted as the additional level of protection against severe disease gained through infection or vaccination, compared to a fully naive individual.} against hospitalisation. 

\subsubsection*{Parametrisation and Calibration} 
Besides input timelines of vaccinations and daily new reported cases the model is parameterised with various assumptions about the immunisation process, in specific using distributions and distribution parameters for immunity waning. So far, we estimated the parameters by fitting survival curves to published data about vaccine effectiveness controlled for the time since vaccination. Besides, the model requires a feasible assumption for the case-detection rate. All other model parameters have a smaller impact on the immunisation level and can be estimated easier. 

\subsubsection*{Implementation and Source Code}
The model was implemented in Python3. The source code to the model implemented in Python3 including a base-parametrisation is found in \href{https://github.com/dwhGmbH/covid19_model_family}{https://github.com/dwhGmbH/covid19\_model\_family}.

\subsection*{\hospitalmodelLong}
\begin{figure}
\centering
\begin{tikzpicture}[-Latex,auto,node distance =1 cm and 1 cm,semithick,
state/.style args={#1}{rounded rectangle, draw, minimum width = 2 cm,fill=#1!10,text=#1,draw=#1, align = center},
edgecolor/.style args={#1}{color=#1},
point/.style = {circle, draw, inner sep=0.04cm,fill,node contents={}},
bidirected/.style={Latex-Latex,dashed},
el/.style = {inner sep=2pt, align = left, sloped},
every edge/.append style={nodes={pos=0.25,anchor=center, circle, draw,fill=white,font = \tiny, inner sep=0.04cm}}]
\input{clds/colordefs}
\def \diC {inputColor};
\def \uiC {ignoreColor};
\def \testsC {ignoreColor};
\def \varC {ignoreColor};
\def \seasC {ignoreColor};
\def \contC {ignoreColor};
\def \polC {ignoreColor};
\def \infC {ignoreColor};
\def \infedC {varColor};
\def \recC {ignoreColor};
\def \immC {ignoreColor};
\def \susC {ignoreColor};
\def \vaccC {ignoreColor};
\def \lostC {ignoreColor};
\def \hospC {varColor};
\def \deadC {ignoreColor};
\def \icuC {outputColor};
\def \nonicuC {outputColor};
\def \transicuC {varColor};
\def \transnonC {varColor};
\def \vartwoC {ignoreColor};
\def \varthreeC {inputColor}

\def \uiinfedC {ignoreColor};
\def \diinfedC {linkColor};
\def \transuiC {ignoreColor};
\def \transdiC {ignoreColor};
\def \seasuiC {ignoreColor};
\def \seasdiC {ignoreColor};
\def \testsdiC {ignoreColor};
\def \testsuiC {ignoreColor};
\def \contuiC {ignoreColor};
\def \contdiC {ignoreColor};
\def \polcontC {ignoreColor};
\def \suscontC {ignoreColor};
\def \immcontC {ignoreColor};
\def \infcontC {ignoreColor};
\def \vaccsusC {ignoreColor};
\def \vaccimmC {ignoreColor};
\def \lostsusC {ignoreColor};
\def \immlostC {ignoreColor};
\def \immsusC {ignoreColor};
\def \recimmC {ignoreColor};
\def \infedrecC {ignoreColor};
\def \infedinfC {ignoreColor};
\def \infedhospC {linkColor};
\def \infeddeadC {ignoreColor};
\def \hospdeadC {ignoreColor};
\def \hospicuC {linkColor};
\def \hospnonicuC {linkColor};
\def \nonicutransicuC {linkColor};
\def \transnonnonicuC {linkColor};
\def \icutransnonC {linkColor};
\def \transicuicuC {linkColor};
\def \transicunonicuC {linkColor};
\def \transnonicuC {linkColor};
\def \varuiC {ignoreColor};
\def \vardiC {ignoreColor};
\def \vartwolostC {ignoreColor};
\def \vartwoimmC {ignoreColor};
\def \varthreedeadC {ignoreColor};
\def \varthreehospC {linkColor};

\input{clds/cld_base}
\end{tikzpicture}
\caption{\cldcaption{\hospitalmodelLong}}
\label{fig:cld_hospmodel}
\end{figure}
Hospital and intensive care unit (ICU) bed occupancy drove Austria's COVID policies in the first two years of the pandemic. Overcrowded hospitals posed as the key argument for policies like quarantine regulations, mandatory face-mask wearing, school closures, and lockdown. To advise decision-makers, we needed to provide projections for these variables. 

Initially, hospitalisations were integrated into the \abm, but its complexity and long run times made it difficult to calibrate. So, we created a separate, simpler stock-flow model called the \hospitalmodelLong (henceforth \hospitalmodel).

This model was developed by the Gesundheit Österreich GmbH in cooperation with the members of the Austrian COVID Forecasting Consortium and uses real and/or predicted reported case numbers as input and provides estimates for the occupancy. It was introduced in \cite{bicher2022supporting} and has since been modified. Here, we introduce a more flexible and generic version of the model, which is better suited for long-term analysis. See Figure \ref{fig:cld_hospmodel} for a causal map of the model.

\subsubsection*{Short Model Description}
The model makes use of the time-series of daily new confirmed cases and maps it onto a time series for the hospital occupancy. It uses a scalar hospitalisation rate and two duration distributions which state (a) how much time passes between positive test and hospitalisation and (b) how long persons stay in the hospital. The model can be regarded as a deterministic difference equation model involving discrete convolutions with the duration distributions:
\begin{align}
\text{admissions}_{(i\rightarrow i+k)}&=\text{cases}_i\cdot \text{rate}\cdot \text{(distribution admissions)}_{k}\\
\text{admissions}_i&=\sum_{k=1}^{i}\text{admissions}_{(k\rightarrow i)}\\
\text{releases}_{(i\rightarrow i+k)}&=\text{admissions}_i\cdot \text{(distribution releases)}_{k}\\
\text{releases}_i&=\sum_{k=1}^{i}\text{releases}_{(k\rightarrow i)}\\
    \text{occupancy}_{i+1}&=\sum_{k=1}^{i}\text{admissions}_k-\text{releases}_k.
\end{align}

A detailed model specification is found in \ref{sec:spec_hospmodel}.

\subsubsection*{Model Usage}
As a deterministic difference equation model, it can be executed highly efficient. In the typical case, the model is used on a concatenated input time-series consisting of reported daily new SARS-CoV-2 cases and a case forecast. It then produces forecasts for the occupancy of both normal beds and ICU beds. For long term forecasts, we usually vary the base hospitalisation rate by including additional assumptions for immunity or virulence dynamics using a second input time-series. This makes the result more feasible and provides a better picture of the uncertainty of the result.


\subsubsection*{Parametrisation and Calibration}
The model is calibrated using historic data of new confirmed daily cases and hospital occupancy. Usually, the most recent 120 days are regarded, where the first 100 days are used as a transient phase and the latter 20 as calibration window (see \ref{sec:spec_hospmodel} for a more detailed description of the calibration process). Note that flattening is usually not necessary, since the performed convolutions by the model provide a rather smooth solution anyway. 

When writing the two duration distributions as functions of the scalar moments of the distribution, standard algorithms like Nelder-Mead simplex can be used as a calibration method.

\subsubsection*{Implementation and Source Code}
The model is implemented in Python using vector operations. The packages \textit{Numpy} and \textit{Scipy} provide routines to make this highly efficient. Finally, the Nelder-Mead simplex implementation from the \textit{Scipy's optimize} package is used to find the optimal parameter set. Full calibration and subsequent simulation only take a few seconds on a standard notebook. Consequently, also hyper-parameter studies, e.g. for different shapes of distributions or different calibration time-frames are possible. The source code to the model including sample input data is found in \href{https://github.com/dwhGmbH/covid19_model_family}{https://github.com/dwhGmbH/covid19\_model\_family}.

\subsection*{\agemodelLong}
\begin{figure}
\centering
\begin{tikzpicture}[-Latex,auto,node distance =1 cm and 1 cm,semithick,
state/.style args={#1}{rounded rectangle, draw, minimum width = 2 cm,fill=#1!10,text=#1,draw=#1, align = center},
edgecolor/.style args={#1}{color=#1},
point/.style = {circle, draw, inner sep=0.04cm,fill,node contents={}},
bidirected/.style={Latex-Latex,dashed},
el/.style = {inner sep=2pt, align = left, sloped},
every edge/.append style={nodes={pos=0.25,anchor=center, circle, draw,fill=white,font = \tiny, inner sep=0.04cm}}]
\input{clds/colordefs}
\def \diC {outputColor};
\def \uiC {ignoreColor};
\def \testsC {ignoreColor};
\def \varC {ignoreColor};
\def \seasC {ignoreColor};
\def \contC {varColor};
\def \polC {ignoreColor};
\def \infC {ignoreColor};
\def \infedC {varColor};
\def \recC {inputColor};
\def \immC {varColor};
\def \susC {varColor};
\def \vaccC {inputColor};
\def \lostC {ignoreColor};
\def \hospC {ignoreColor};
\def \deadC {ignoreColor};
\def \icuC {ignoreColor};
\def \nonicuC {ignoreColor};
\def \transicuC {ignoreColor};
\def \transnonC {ignoreColor};
\def \vartwoC {ignoreColor};
\def \varthreeC {ignoreColor}

\def \uiinfedC {ignoreColor};
\def \diinfedC {linkColor};
\def \transuiC {ignoreColor};
\def \transdiC {ignoreColor};
\def \seasuiC {ignoreColor};
\def \seasdiC {ignoreColor};
\def \testsdiC {ignoreColor};
\def \testsuiC {ignoreColor};
\def \contuiC {ignoreColor};
\def \contdiC {linkColor};
\def \polcontC {ignoreColor};
\def \suscontC {linkColor};
\def \immcontC {linkColor};
\def \infcontC {ignoreColor};
\def \vaccsusC {linkColor};
\def \vaccimmC {linkColor};
\def \lostsusC {ignoreColor};
\def \immlostC {ignoreColor};
\def \immsusC {linkColor};
\def \recimmC {linkColor};
\def \infedrecC {linkColor};
\def \infedinfC {ignoreColor};
\def \infedhospC {ignoreColor};
\def \infeddeadC {ignoreColor};
\def \hospdeadC {ignoreColor};
\def \hospicuC {ignoreColor};
\def \hospnonicuC {ignoreColor};
\def \nonicutransicuC {ignoreColor};
\def \transnonnonicuC {ignoreColor};
\def \icutransnonC {ignoreColor};
\def \transicuicuC {ignoreColor};
\def \transicunonicuC {ignoreColor};
\def \transnonicuC {ignoreColor};
\def \varuiC {ignoreColor};
\def \vardiC {ignoreColor};
\def \vartwolostC {ignoreColor};
\def \vartwoimmC {ignoreColor};
\def \varthreedeadC {ignoreColor};
\def \varthreehospC {ignoreColor};

\input{clds/cld_base}

\path (infed) edge[bend left=27,edgecolor=linkColor,dashed] node {$+$}  (cont);

\node[state=inputColor,below of =seas] (beta) {$\beta$};
\path (beta) edge[bend right=20,edgecolor=linkColor,dashed] node {$+$}  (cont);
\end{tikzpicture}
\caption{\cldcaption{\agemodelLong}}
\label{fig:cld_agemodel}
\end{figure}
The age structure of infected individuals, split by vaccination status, is a crucial input for the \hospitalmodel. For short-term forecasts, the current distribution can be extrapolated. However, this strategy is not viable for medium- or long-term scenario simulations. 

Although the \abm can be used to evaluate disease waves with respect to age structure, it is challenging to calibrate the model for the current age distribution of cases. This is because, like most other SEIR-type models, simulations cannot be simply started at an arbitrary point in time (see \cite{bicher2021evaluation}). Since many events from the past impact the dynamics of the near future, simulations always have to be started from the very beginning of the pandemic.

To overcome this problem, the \agemodelLong (\agemodel) was developed. By neglecting the "exposed"-state of an infected person and limiting the model structure to a SIR-type, the model became ``memoryless'' in the sense that it can be initialised with observed data (active cases, vaccinated cases, etc.). While the model's epidemiological accuracy for forecasting case numbers may suffer from this simplification, the dynamics of the age structure of the cases is well predicted.

\subsubsection*{Short Model Description}
Motivated by a work of A.G. McKendrick ~\cite{mkendrick_applications_1925}, which he published the year before his groundbreaking publication about the concept of susceptible -- infectious -- recovered (SIR) modelling together with W. Kermack~\cite{kermack_contribution_1927}, we decided to develop an epidemic compartment model wherein age is a second continuous variable next to time. For example, the compartment of susceptible individuals $S=S(a,t)$ is analysed as a function of time $t$ and age $a$. This way, the approach essentially becomes a partial differential equation (PDE). Key feature of the model is a contact kernel which decides about contact between infectious persons with age $a_1$ with susceptible persons with age $a_2$. Models following this strategy are well known and their properties are well analysed (see \cite{hoppensteadt_age_1974,dietz_proportionate_1985, bicher2017comparison}). Our approach founds on a classic SIR model published in \cite{hoppensteadt_age_1974} and was extended by a second disease path to depict vaccinations and vaccine effectiveness. For more information the reader is referred to the detailed model specification (\ref{sec:spec_agemodel}).

\subsubsection*{Model Usage}
Although the \agemodel itself is an epidemiological model, its main purpose is not forecasting of disease numbers. For this purpose the contact process, the disease path and the immunisation process are too much simplified. It is purely used to investigate the dynamics of the age-distribution of infected persons. 
An age-dependent contact kernel and age-dependent information on previous infections and vaccinations are used as model input. Usually the model is then calibrated to a given disease progression over the course of an epidemic wave to provide information about the current age-distribution among the infected cases. Hence, the detected infections pose both input via the overall number as calibration reference, as well as variable of interest via their age structure.

\subsubsection*{Parametrisation and Calibration}
Age-dependent surveillance data about previous infections and vaccinations are evaluated to provide feasible initial conditions. One of the most valuable features of the model is that it is well capable of being initialised by data with different age resolutions. This is guaranteed by a kernel density estimation (KDE) performed on top of the data-sets. This KDE is required anyway to make the initial curves differentiable.
The age-dependent contact kernel is the key parameter of the model. Thanks to fantastic studies like POLYMOD \cite{mossong2017polymod} or COMIX \cite{coletti2020comix} there is lots of public data available on this subject.

The calibration to a specific disease progression is done by varying the parameters of the time dependent infectiousness parameter function $\beta$ which can be interpreted as a summary of policies, seasonality and infectiousness of the virus (variant). In principle there is no limitation on where the calibration reference comes from. Typically, either historic data from previous disease waves or forecasts from other more accurate models such as the \abm are used. Calibration is performed with an iterative bisection method.





\subsubsection*{Implementation and Source Code}
Due to its great numerical properties, the model is developed in MATLAB. To solve the PDE a standard Method of Lines approach is chosen with a Numerical Differential Formula time-integrator (MATLAB's ode15s solver\cite{ode15s}). The integral parts on the right-hand side of the equation (see \ref{eq:contacts}) are solved using the trapezoid-method. The source code of the model can be found in \href{https://github.com/dwhGmbH/covid19_model_family}{https://github.com/dwhGmbH/covid19\_model\_family}. 

\section*{Results}
This work emphasizes the benefits of using a model family, and therefore our interpretation of a result differs from classical modelling and simulation studies. We will not delve into specific simulation outcomes, but rather focus on how the result was generated and used in decision support. Four success stories will illustrate how the models created value in decision making, and representative model outcomes will be presented to demonstrate this value. For the result figures displayed in this work, open data interfaces of the Austrian Ministry of Health and the Austrian Agency for Health and Food Safety GmbH (AGES) were used to gather the parametrization/input data for daily new confirmed cases, variant distribution, vaccination rates and hospital occupancy.

\paragraph{\sst{1}: Combined usage of the \abm and the \hospitalmodel}
In April 2020, about a month after the first detected case of SARS-COV-2 in Austria, the COVID Forecasting Consortium of the Ministry of Health was established. By 2023, the consortium produced and published more than 150 short-term forecasts of SARS-CoV-2 case numbers and COVID-19 hospital bed occupancy (see \href{https://datenplattform-covid.goeg.at/prognosen}{https://datenplattform-covid.goeg.at/prognosen}). Forecast generation involved three modelling groups each producing a case number forecast using an epidemiological model. The TU Wien used the \abm, while the other two groups used macroscopic modelling approaches. These forecasts of detected infections were then harmonised into an ensemble forecast which was used as input for a common occupancy model producing the final forecast for the hospital occupancy. The \hospitalmodel from Section \ref{sec:spec_hospmodel} is a simplified version of this pavement model (see Section \ref{sec:spec_hospmodel}).

The splitting of the forecasting process into caseload and pavement forecasting should prove to be one of the cornerstones of the consortium's success. The strategy helps validate, verify, and compare the epidemiological models, some of which are highly complex. It simplifies and accelerates scenario calculations by allowing for uncertainties and different assumptions in both forecast sections. Finally, it is also flexible, fail-safe, and the results are easily reproducible.

\begin{figure}
    \centering
    \includegraphics[width=\textwidth]{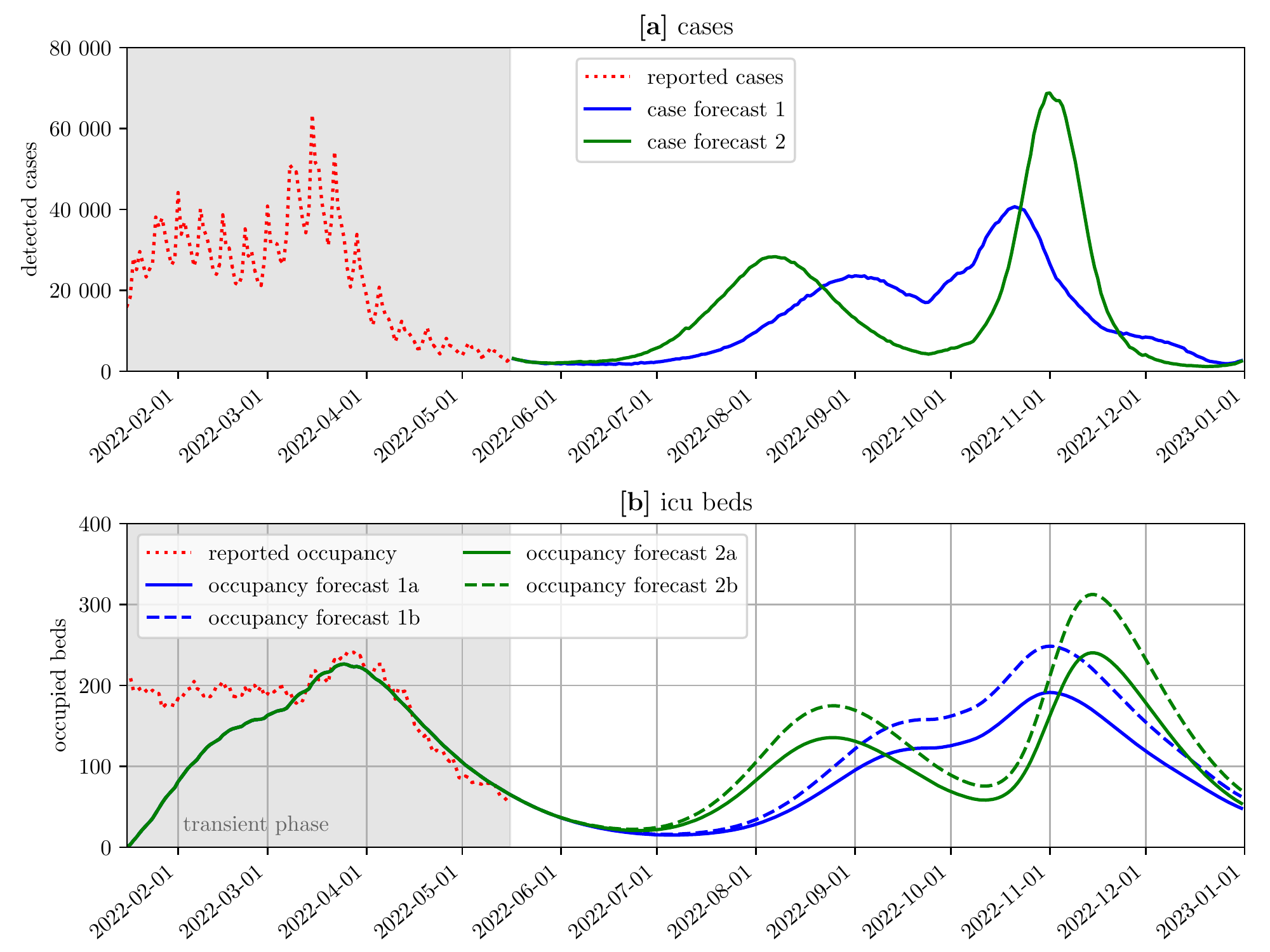}
    \caption{Combining different case forecast scenarios from the \abm [a] with different assumptions for virulence (equal and $30\%$ increased) of the a new variant in the \hospitalmodel [b].}
    \label{fig:ss1}
\end{figure}

Figure \ref{fig:ss1} provides an example of forecasts generated by combining the two models. They were produced on 2022-05-16 to estimate the potential burden on the hospitals resulting from the emergence of the new variant $BA_{4/5}$. Assumptions about the higher infectivity and immune escape of the variant were handled by the \abm which thus produced different forecasts for the case numbers. Further assumptions on the virulence were included in the \hospitalmodel which thus generated different forecast for the hospital occupancy for each of the result scenario of the \abm. These forecasts provided an early image of the possible range of hospital occupancy. As soon as better information for properties of the new variant was available in the literature, the range of the results could be narrowed down.



\paragraph{\sst{2}: Combined usage of the \agemodel and epidemiological models}
In August 2021, the Austrian COVID Forecasting Consortium (see \cite{bicher2022supporting}) was tasked to summarise findings and scenario based forecasts for the upcoming Delta wave, with a focus on ICU occupancy. Therefore scenario-based forecasts for reported cases were developed and the \hospitalmodel was applied to translate cases to occupancy. Although this strategy has been well applicable for the prior disease waves, vaccinated persons needed to be factored into the computations now -- about $56\%$, primarily elderly, have received at least two vaccination doses by September 2021\cite{herbek2012electronic}. Vaccinated persons were already known to have a reduced infection, but even lower hospitalisation and ICU risk. In order to quantify this advantage, estimates were needed how the case population would be split into age \underline{and} vaccination groups.

\begin{figure}
    \centering
    \includegraphics[width=\textwidth]{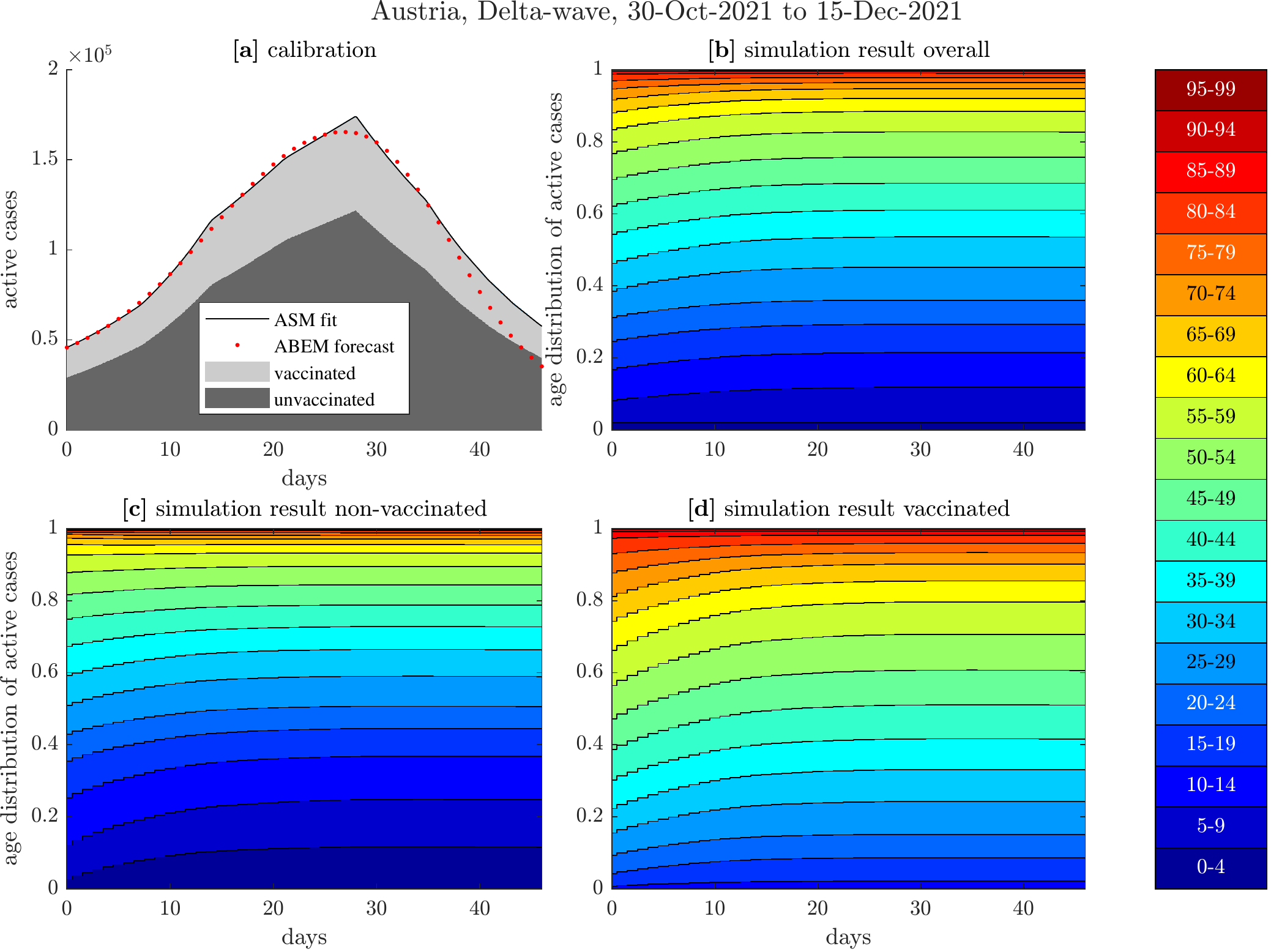}
    \caption{Result of the \agemodel when fitted to a forecast from the \abm between 30-Oct-2021 to 15-Dec-2021 (Delta wave). Part [a] shows reference data, the fitted simulation result and the split between vaccinated and unvaccinated cases. Part [b] shows age distributions of all active cases. Parts [c] and [d] show the age distribution of active cases separately for people with and without vaccination.}
    \label{fig:ss3_a}
\end{figure}

Therefore, the \agemodel was applied in addition to the \abm. First, a medium-range forecast for the Delta wave was generated with the \abm. Then, the \agemodel was initialised to the current population distribution with respect to vaccinated, recovered and infected persons. Finally, the \agemodel was calibrated to match the case number forecast generated by the \abm. This was done by using a step-function $\beta(a,t):=\hat{\beta}(a)\sum_{i=0}^{6}1_{[7i,7i+7)}(t)\beta_i$ for the transmissibility, and by fitting the seven scalar parameters $\beta_0,\dots,\beta_6$ (we refer to the model specification for details). The results of the \agemodel provided a proper insight into the expected age distribution of overall, vaccinated and non-vaccinated cases in the upcoming wave and are shown in Figure \ref{fig:ss3_a}.

\begin{figure}
    \centering
    \includegraphics[width=\textwidth]{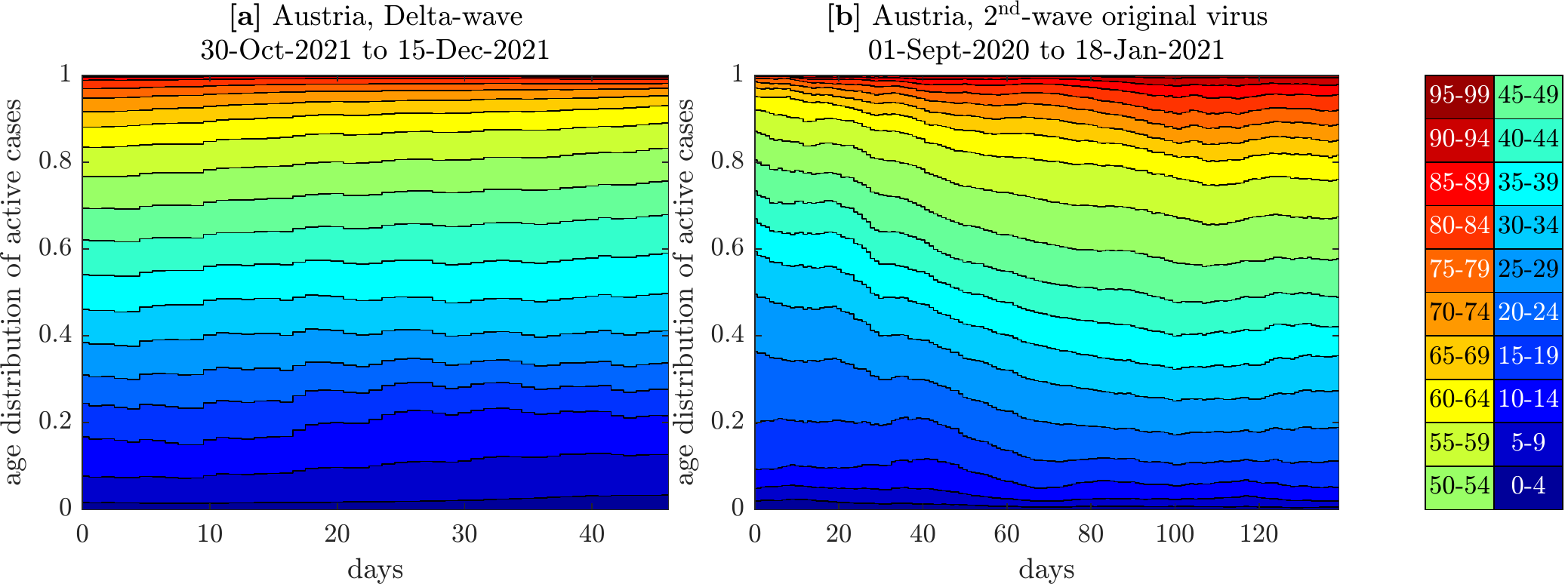}
    \caption{Comparison of the reported age distribution of active cases between the a in autumn 2021 [a] and a wave in autumn 2020 [b]. Because mainly older people were vaccinated, the 2021 delta-wave showed clearly different age dynamics than the other waves.}
    \label{fig:ss3_b}
\end{figure}

Model results (correctly) showed that the Delta wave shifted the active cases towards younger age cohorts, a result of older age groups being prioritised in the vaccination program, leaving many children insufficiently vaccinated by autumn 2021. The age distribution of vaccinated and non-vaccinated cases (lower two plots in Figure \ref{fig:ss3_a}) illustrate this problem. This stood in contrast to previous waves, which were initialised with younger cohorts and shifted towards the older ones during the upswing of case numbers. (see Figure \ref{fig:ss3_b}). Up to the current date, the Delta wave was the only one showing this very profile.

Results had important implications for the expected hospitalisation rates. Despite the majority of cases being non-vaccinated, their relatively young age profile indicated a low hospitalisation risk, with hospitalisation rates less than half the size of rates in a fully non-vaccinated population. Limits for critical ICU occupancy (33\%) were increased from about 2400 to 5100 daily new confirmed cases in the steady state. Results were published on the homepage of the Austrian Ministry of Health, see \cite{policy_letter_autumn}.



\paragraph{\sst{3}: Combined usage of \abm and \waningmodel.} Since March 2021, when Austria had already observed two epidemic waves and the vaccine started to become widely available, the \waningmodel was used in combination with case data from the official reporting system to generate monthly estimates for the level of immunity against certain targets, typically infection with a specific variant or severe disease progression. Results help to get an image of the current pandemic risk and were published monthly on \href{http://www.dexhelpp.at/en/immunization_level}{http://www.dexhelpp.at/en/immunization\_level}. 

\begin{figure}
    \centering
    \includegraphics[width=\textwidth]{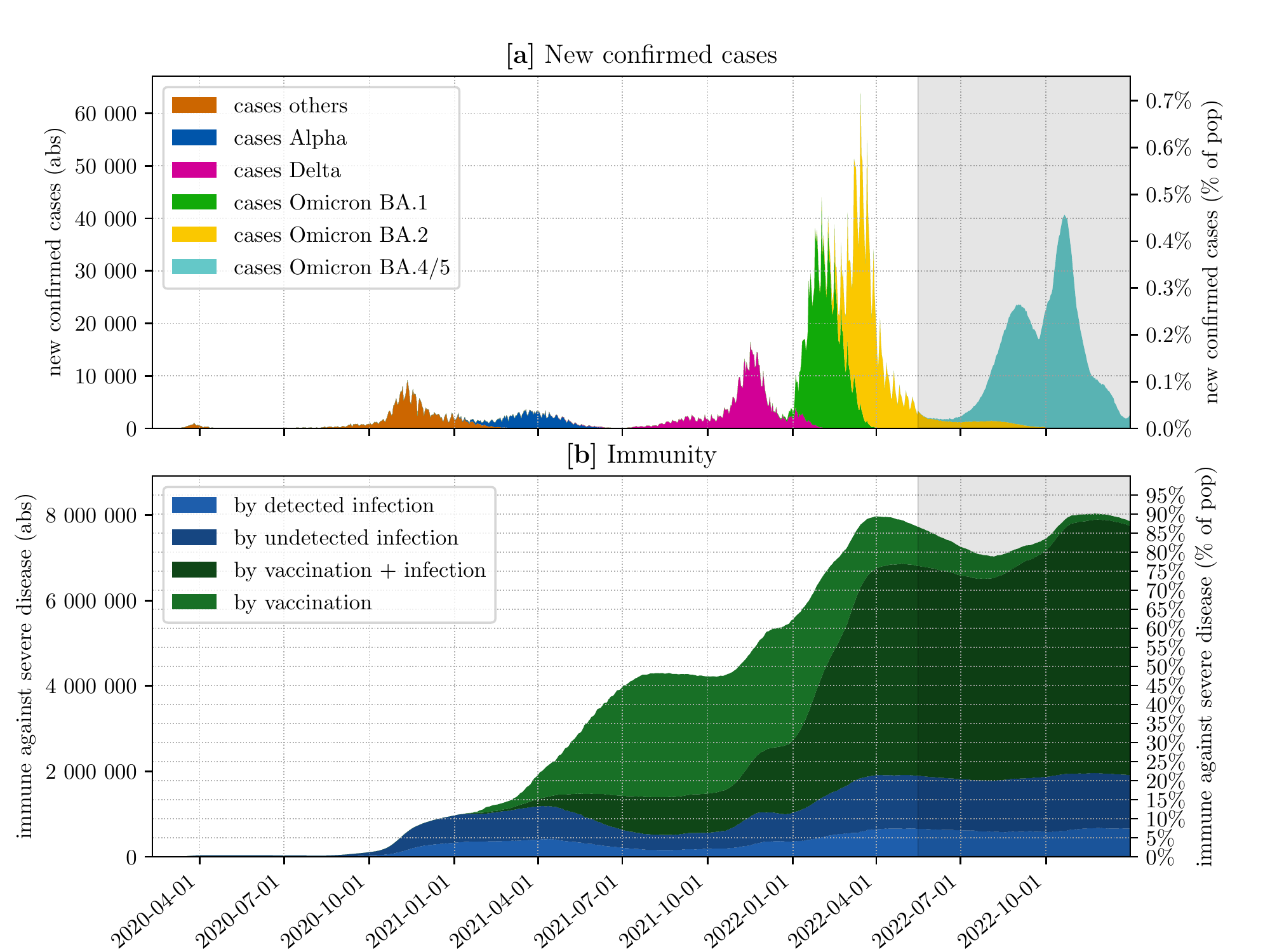}
    \caption{Combined usage of \waningmodel and \abm. The variant specific case numbers [a] are fed into the \waningmodel to estimate the level of immunity [b] against severe disease progression (i.e. hospitalisation). Combined with a case number forecast from the \abm (grey area), a forecast for the immunity level can be made. The shown forecast was based on data until May 16th, 2022.}
    \label{fig:ss2}
\end{figure}

An example of this model application is shown in Figure \ref{fig:ss2}. Case data up to May 16th, 2022 (part [a]) has been fed into the \waningmodel to estimate the time dynamics of the immunity level of the population against severe disease progression (part [b]). The historical case data has then been extended with a forecast for the future dynamics of the new variant Omicron BA.4/5 generated using the \abm (grey area in part [a]). Applying the \waningmodel on the joint time-series of case data and forecast, a prognosis for the immunity level was made, seen in the grey area of part [b] of the figure.

These results provided valuable insights to the decision makers since they gave a proper image of possible but also impossible long-term strategies to overcome the COVID-19 crisis. For details and interpretation of the specific results we refer to  \href{https://www.dwh.at/en/news/covid-19-scenario-simulations-for-summer-autumn-winter-2022/}{https://www.dwh.at/en/news/covid-19-scenario-simulations-for-summer-autumn-winter-2022/}.

\paragraph{\sst{4}: Joint usage of the \abm, \waningmodel and the \hospitalmodel}
In the course of the scenario calculations  on the future dynamics of infections driven by Omikron.BA.4/5 (see \sst{3} and \cite{policy_letter_autumn}, respectively), the potential impact on the utilisation of Austrian hospitals was also evaluated. For this purpose, both the case number scenarios of \abm and the corresponding immunity levels from \waningmodel were used as input to the \hospitalmodel. The latter idea is based on the basic assumption that the hospitalisation rate is directly proportional to the proportion of the risk group among those infected. Defining the risk group as the proportion of those susceptible to infection who are not protected against a severe disease progression, this ratio can be calculated from the corresponding result curves of the \waningmodel: Let $PS$ denote being protected against severe disease and $PI$ against infection, then 
\begin{equation}
    \text{hospitalisationrate}\propto P(\neg PS|\neg PI)
    =\frac{P(\neg(PS \vee PI))}{P(\neg PI)}\underbrace{=}_{PI \Rightarrow PS}\frac{P(\neg PS)}{P(\neg PI)}=\frac{1-P(PS)}{1-P(PI)}.
\end{equation}

\begin{figure}
    \centering
    \includegraphics[width=\textwidth]{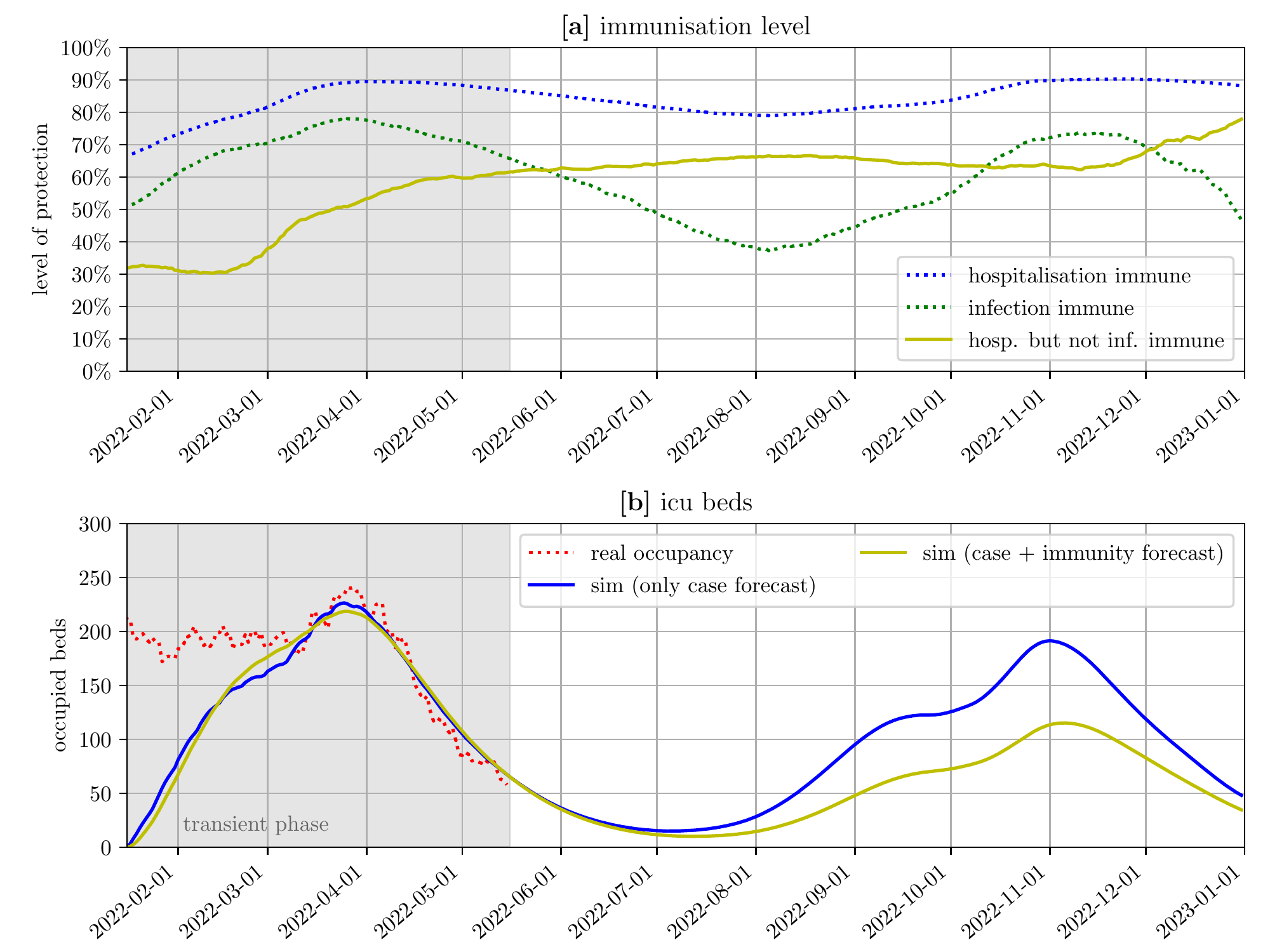}
    \caption{Comparison of hospital forecast with and without regarding the immunisation level against severe disease. Part [a] shows the dynamics of the different levels of protection estimated with the \waningmodel. The dotted lines represent the probability to be immune against infection and severe disease progression (hospitalisation) respectively. The yellow line shows the conditional probability to be protected against severe disease if one is not protected against infection. Part [b] shows the effect of either neglecting or including this conditional immunisation as additional input to the \hospitalmodel.}
    \label{fig:ss4}
\end{figure}

Figure \ref{fig:ss4} shows one result from this study. Section [a] visualises the dynamics of the different levels of protection. The effect on the \hospitalmodel results can be seen in sections [b] when comparing the blue and yellow curves: fast decreasing protection against infection in the prognosis period increases the relative protection of susceptibles against severe progression and correspondingly decreases the overall hospitalisation rate. 

This observation was one of many that was valuable to decision makers from this analysis. The prospect of hospital loads again reaching similar high ranges in the winter of 2022 than in 2021 also provided added value to planning.

\section{Discussion}
In the present work we described development, specification and usage of four entirely different models describing one part of a large system. Each model comes with different modelling purpose, input, output and limitations due to its view and model boundaries. The four presented success stories are not only examples for the successful joint use of the models, they also highlight the advantages of the model family in contrast to one large, complex stand-alone model:

\textit{Model Resolution and Validity.} Each model in the family is itself stand-alone and can be customised accordingly in the choice of the modelling method and model resolution to fit the problem. For example, \abm and \waningmodel are each microscopic, \hospitalmodel and \agemodel are considered to be macroscopic. Also, \abm and \waningmodel differ greatly in the level of detail of individuals and scalability. The adequate choice of the modelling strategy and resolution are not only basic requirements of general good modelling practices, but are also in advantage to the large stand-alone model, where the resolution of the whole model is fixed by the resolution of the component which requires the highest resolution. This circumstance is often problematic, as demonstrated in \sst{2}. Due to its high resolution and high sensitivity the \abm is not well suited for simulating age shift in infection waves. However, in the model family it can be supported by the much lower resolution \agemodel.

\textit{Computation time and parametrisation efforts.} Because (a) from the model family only those models are used, which are necessary for the respective problem, and (b) the resolution of the individual models is usually lower than the one of the stand-alone model, the computational effort for experiments in the model family is usually lower. The same is true for the parametrisation effort and the potential sources of error. This advantage was exploited in \sst{3}: Even though in principle all scenarios could have been computed with \abm alone, the sequential use of \abm and \waningmodel greatly reduced the computational effort, the error-sensitivity and the effort for parametrisation. 

\textit{Flexibility.} The interfaces between the models, i.e. input and output, offer many possibilities for manual or automated intervention in the process. This makes the models essentially modules, which can be used even beyond the boundaries of the research institution. \sst{1} demonstrates this on the example of the ensemble forecast, which is used as input to the pavement model.

\textit{Validation and verification.} Each of the models in the family can be independently validated and verified. Model uncertainties and parameter sensitivities can be determined individually. Accordingly, experiments based on the linkage of the models are as valid and correct as the individual models. The uncertainty of the result can be derived from the uncertainties of the individual models. To perform an appropriate model analysis for a large stand-alone model, each component of the model would have to be analysed with the same care as the corresponding model of the model family. However, this would become much more costly with the longer computation time and higher parametrisation effort of the stand-alone model. As additional bonus, the overlap regions of the individual models can be used for cross-model validation (see \cite{popper2015comparative}).

\textit{Communication.} In contrast to the stand-alone model, the model family provides a clear structure for communicating models and model results. They can be communicated individually and therefore do not have to be immediately understood as a whole.

\textit{Efficient creation and use.} The model family is also advantageous from a project management perspective. Implementation, maintenance, extension, analysis, execution, etc. can be distributed (and passed on) much better to several persons or project teams. Thus progress can be made much more efficient. In the contrast, splitting simulation experiments on multiple models requires a whole and well understood picture of strengths, weaknesses, in-, output and boundaries of the individual models. In our applications, we found the causal loop diagrams helpful to get a quick overview and proper assignment of the given tasks to the right model(s).

Even apart from its value in collaborative work, the usage of (modified) causal loop diagrams has proven useful for representing and comparing the individual models. First, superimposing the diagrams provides a complete picture of the processes mapped in the family. One can see overlapping areas, which can be used for cross-validation, input-output relationships between the models, which would allow sequential simulation, and poorly covered areas, which indicate weaknesses in the family and can serve as motivation to create new models. The diagram of a single model immediately shows neglected causal relationships and broken causal loops, which can be useful for validation.

Undoubtedly, development, implementation, parametrisation, validation and usage of a whole model family came with huge efforts and time expenses. Nevertheless, our experience in working with these models for over two years showed, that development of distinct models for specific subsystems essentially saved time in the long run, due to the increased stability and smaller run-times of less complex models. Summarising, we clearly recommend modellers working on decision support in large and complex systems to invest time for the development of model families instead of one large, complex stand-alone model. Maintaining two or more models in parallel causes overheads, but pays off in the long run. In this approach, it does not matter, whether a family was planned right from the start or existing models are split as soon as they become too large and complex. In the latter case, a causal loop diagram of the overall system can be useful to determine which links can be neglected or which feedback loops can be broken without causing additional model errors.

\section*{Acknowledgements}
We thank the Gesundheit Österreich GmbH (GÖG) and the Complexity Science Hub Vienna (CSH) for their great collaboration in the time of the COVID-19 crisis and the Austrian Ministry of Health and the Austrian Agency for Health and Food Safety GmbH (AGES) for providing various open COVID-19 related data.

\bibliographystyle{plain}
\bibliography{references}

\begin{thebibliography}{10}

\bibitem{Amman_Viral_2022}
Fabian Amman, Rudolf Markt, Lukas Endler, Sebastian Hupfauf, Benedikt Agerer,
  Anna Schedl, Lukas Richter, Melanie Zechmeister, Martin Bicher, Georg Heiler,
  Petr Triska, Matthew Thornton, Thomas Penz, Martin Senekowitsch, Jan Laine,
  Zsofia Keszei, Peter Klimek, Fabiana Nägele, Markus Mayr, Beatrice Daleiden,
  Martin Steinlechner, Harald Niederstätter, Petra Heidinger, Wolfgang Rauch,
  Christoph Scheffknecht, Gunther Vogl, Günther Weichlinger, Andreas~Otto
  Wagner, Katarzyna Slipko, Amandine Masseron, Elena Radu, Franz Allerberger,
  Niki Popper, Christoph Bock, Daniela Schmid, Herbert Oberacher, Norbert
  Kreuzinger, Heribert Insam, and Andreas Bergthaler.
\newblock Viral variant-resolved wastewater surveillance of sars-cov-2 at
  national scale.
\newblock {\em Nature Biotechnology}, 40(12):1814–--1822, Jul 2022.

\bibitem{balaban2015toward}
Mariusz~A Balaban.
\newblock {\em Toward a theory of multi-method modeling and simulation
  approach}.
\newblock Old Dominion University, 2015.

\bibitem{bicher2017comparison}
M~Bicher, N~Popper, and G~Schneckenreither.
\newblock Comparison of a microscopic and a macroscopic age-dependent sir
  model.
\newblock {\em Mathematical and Computer Modelling of Dynamical Systems},
  23(2):177--195, 2017.

\bibitem{bicher2022time}
Martin Bicher, Claire Rippinger, and Niki Popper.
\newblock Time dynamics of the spread of virus mutants with increased
  infectiousness in austria.
\newblock {\em Ifac-papersonline}, 55(20):445--450, 2022.

\bibitem{bicher2022model}
Martin Bicher, Claire Rippinger, G{\"u}nter Schneckenreither, Nadine Weibrecht,
  Christoph Urach, Melanie Zechmeister, Dominik Brunmeir, Wolfgang Huf, and
  Niki Popper.
\newblock Model based estimation of the sars-cov-2 immunization level in
  austria and consequences for herd immunity effects.
\newblock {\em Scientific Reports}, 12(1):1--15, 2022.

\bibitem{Bicher_Evaluation_2021}
Martin Bicher, Claire Rippinger, Christoph Urach, Dominik Brunmeir, Uwe
  Siebert, and Niki Popper.
\newblock Evaluation of contact-tracing policies against the spread of
  sars-cov-2 in austria: An agent-based simulation.
\newblock {\em Medical Decision Making}, 41(8):1017--–1032, May 2021.

\bibitem{bicher2021evaluation}
Martin Bicher, Claire Rippinger, Christoph Urach, Dominik Brunmeir, Uwe
  Siebert, and Niki Popper.
\newblock Evaluation of contact-tracing policies against the spread of
  sars-cov-2 in austria: An agent-based simulation.
\newblock {\em Medical Decision Making}, 41(8):1017--1032, 2021.
\newblock PMID: 34027734.

\bibitem{Bicher_Iterative_2022}
Martin Bicher, Claire Rippinger, Melanie Zechmeister, Beate Jahn, Gaby
  Sroczynski, Nikolai Mühlberger, Julia Santamaria-Navarro, Christoph Urach,
  Dominik Brunmeir, Uwe Siebert, and Niki Popper.
\newblock An iterative algorithm for optimizing covid-19 vaccination strategies
  considering unknown supply.
\newblock {\em PLOS ONE}, 17(5):e0265957, May 2022.

\bibitem{bicher2018gepoc}
Martin Bicher, Christoph Urach, and Niki Popper.
\newblock Gepoc abm: a generic agent-based population model for austria.
\newblock In {\em 2018 Winter Simulation Conference (WSC)}, pages 2656--2667.
  IEEE, 2018.

\bibitem{bicher2022supporting}
Martin Bicher, Martin Zuba, Lukas Rainer, Florian Bachner, Claire Rippinger,
  Herwig Ostermann, Nikolas Popper, Stefan Thurner, and Peter Klimek.
\newblock Supporting covid-19 policy-making with a predictive epidemiological
  multi-model warning system.
\newblock {\em Communications Medicine}, 2(1):157, 2022.

\bibitem{BRAILSFORD2019721}
Sally~C. Brailsford, Tillal Eldabi, Martin Kunc, Navonil Mustafee, and
  Andres~F. Osorio.
\newblock Hybrid simulation modelling in operational research: A
  state-of-the-art review.
\newblock {\em European Journal of Operational Research}, 278(3):721--737,
  2019.

\bibitem{brauer2008compartmental}
Fred Brauer.
\newblock {\em Compartmental models in epidemiology}, pages 19--79.
\newblock Springer, 2008.

\bibitem{coletti2020comix}
Pietro Coletti, James Wambua, Amy Gimma, Lander Willem, Sarah Vercruysse, Bieke
  Vanhoutte, Christopher~I Jarvis, Kevin Van~Zandvoort, John Edmunds, Philippe
  Beutels, et~al.
\newblock Comix: comparing mixing patterns in the belgian population during and
  after lockdown.
\newblock {\em Scientific reports}, 10(1):1--10, 2020.

\bibitem{policy_letter_autumn}
{COVID Prognose Konsortium}.
\newblock Update des policy briefs vom 8.7.2021 - aktualisierung der
  risikobewertung, szenarien und handlungsanleitungen für den herbst 2021.
\newblock
  \url{https://www.sozialministerium.at/dam/jcr:8847f88c-b314-4d86-9d2b-3f169e047b0e/Policy_Brief_Update_20210831.pdf},
  2021.

\bibitem{crout2008chapter}
N~Crout, Teemu Kokkonen, AJ~Jakeman, JP~Norton, LTH Newham, R~Anderson,
  H~Assaf, BFW Croke, N~Gaber, J~Gibbons, et~al.
\newblock Chapter two good modelling practice.
\newblock {\em Developments in Integrated Environmental Assessment}, 3:15--31,
  2008.

\bibitem{R-4252-DARPA}
Paul~K. Davis.
\newblock {\em An Introduction to Variable-Resolution Modeling and
  Cross-Resolution Model Connection}.
\newblock RAND Corporation, Santa Monica, CA, 1993.

\bibitem{MR-1004-DARPA}
Paul~K. Davis and James~H. Bigelow.
\newblock {\em Experiments In Multiresolution Modeling (MRM)}.
\newblock RAND Corporation, Santa Monica, CA, 1998.

\bibitem{dietz_proportionate_1985}
Klaus Dietz and Dieter Schenzle.
\newblock Proportionate mixing models for age-dependent infection transmission.
\newblock {\em Journal of Mathematical Biology}, 22(1), June 1985.

\bibitem{abt_hp}
dwh GmbH.
\newblock News entry for the abt simulation framework.
\newblock
  \url{http://www.dwh.at/en/news/the-power-of-the-abt-simulation-framework/},
  2020.
\newblock Accessed: 2020-04-17.

\bibitem{forrester1970urban}
Jay~W Forrester.
\newblock Urban dynamics.
\newblock {\em IMR; Industrial Management Review (pre-1986)}, 11(3):67, 1970.

\bibitem{forrester1987lessons}
Jay~W Forrester.
\newblock Lessons from system dynamics modeling.
\newblock {\em System Dynamics Review}, 3(2):136--149, 1987.

\bibitem{golomb1971mathematical}
Solomon~W Golomb.
\newblock Mathematical models: Uses and limitations.
\newblock {\em IEEE Transactions on Reliability}, 20(3):130--131, 1971.

\bibitem{hafner2017terminology}
Irene Hafner and Niki Popper.
\newblock On the terminology and structuring of co-simulation methods.
\newblock In {\em Proceedings of the 8th International Workshop on
  Equation-Based Object-Oriented Modeling Languages and Tools}, pages 67--76,
  2017.

\bibitem{herbek2012electronic}
Susanne Herbek, HA~Eisl, Martin Hurch, Anton Schator, St~Sabutsch, G{\"u}nter
  Rauchegger, Alexander Kollmann, Th~Philippi, Pia Dragon, Elisabeth Seitz,
  et~al.
\newblock The electronic health record in austria: a strong network between
  health care and patients.
\newblock {\em European Surgery}, 44:155--163, 2012.

\bibitem{hoppensteadt_age_1974}
Frank Hoppensteadt.
\newblock An {Age} {Dependent} {Epidemic} {Model}.
\newblock {\em Journal of the Franklin Institute}, 297(5):325--333, May 1974.

\bibitem{Jahn_Targeted_2021}
Beate Jahn, Gaby Sroczynski, Martin Bicher, Claire Rippinger, Nikolai
  Mühlberger, Júlia Santamaria, Christoph Urach, Michael Schomaker, Igor
  Stojkov, Daniela Schmid, Günter Weiss, Ursula Wiedermann, Monika
  Redlberger-Fritz, Christiane Druml, Mirjam Kretzschmar, Maria Paulke-Korinek,
  Herwig Ostermann, Caroline Czasch, Gottfried Endel, Wolfgang Bock, Nikolas
  Popper, and Uwe Siebert.
\newblock Targeted covid-19 vaccination (tav-covid) considering limited
  vaccination capacities—an agent-based modeling evaluation.
\newblock {\em Vaccines}, 9(5), 2021.

\bibitem{kermack_contribution_1927}
W.~O. Kermack and A.~G. McKendrick.
\newblock A {Contribution} to the {Mathematical} {Theory} of {Epidemics}.
\newblock {\em Proceedings of the Royal Society A: Mathematical, Physical and
  Engineering Sciences}, 115(772):700--721, August 1927.

\bibitem{krijkamp2018microsimulation}
Eline~M Krijkamp, Fernando Alarid-Escudero, Eva~A Enns, Hawre~J Jalal,
  MG~Myriam Hunink, and Petros Pechlivanoglou.
\newblock Microsimulation modeling for health decision sciences using r: a
  tutorial.
\newblock {\em Medical Decision Making}, 38(3):400--422, 2018.

\bibitem{ode15s}
Mathworks.
\newblock ode15s, solve stiff differential equations and daes --- variable
  order method, 2022.
\newblock accessed 2022-05-17.

\bibitem{mkendrick_applications_1925}
A.~G. McKendrick.
\newblock Applications of {Mathematics} to {Medical} {Problems}.
\newblock {\em Proceedings of the Edinburgh Mathematical Society}, 44:98,
  February 1926.

\bibitem{miksch2011pin101}
Florian Miksch, Christoph Urach, Niki Popper, G{\"u}nther Zauner, Gottfried
  Endel, I~Schiller-Fr{\"u}hwirth, and F~Breitenecker.
\newblock Pin101 new insights on the spread of influenza through agent based
  epidemic modeling.
\newblock {\em Value in Health}, 14(7):A284, 2011.

\bibitem{mossong2017polymod}
Jo{\"e}l Mossong, Niel Hens, Mark Jit, Philippe Beutels, Kari Auranen, Rafael
  Mikolajczyk, Marco Massari, Stefania Salmaso, Gianpaolo~Scalia Tomba, Jacco
  Wallinga, et~al.
\newblock Polymod social contact data, 2017.

\bibitem{mossong_polymod_2017}
Joël Mossong, Niel Hens, Mark Jit, Philippe Beutels, Kari Auranen, Rafael
  Mikolajczyk, Marco Massari, Stefania Salmaso, Gianpaolo~Scalia Tomba, Jacco
  Wallinga, and {others}.
\newblock {POLYMOD} social contact data, 2017.

\bibitem{pian1973hybrid}
Theodore~HH Pian.
\newblock Hybrid models.
\newblock In {\em Numerical and computer methods in structural mechanics},
  pages 59--78. Elsevier, 1973.

\bibitem{popper2015comparative}
Nikolas Popper.
\newblock {\em Comparative modelling and simulation: a concept for modular
  modelling and hybrid simulation of complex systems}.
\newblock PhD thesis, TU Wien, 2015.

\bibitem{Popper_Synthetic_2021}
Nikolas Popper, Melanie Zechmeister, Dominik Brunmeir, Claire Rippinger, Nadine
  Weibrecht, Christoph Urach, Martin Bicher, Günter Schneckenreither, and
  Andreas Rauber.
\newblock Synthetic reproduction and augmentation of covid-19 case reporting
  data by agent-based simulation.
\newblock {\em Data Science Journal}, 20:16, Apr 2021.

\bibitem{Rippinger_Evaluation_2021}
C.~Rippinger, M.~Bicher, C.~Urach, D.~Brunmeir, N.~Weibrecht, G.~Zauner,
  G.~Sroczynski, B.~Jahn, N.~Mühlberger, U.~Siebert, and N.~Popper.
\newblock Evaluation of undetected cases during the covid-19 epidemic in
  austria.
\newblock {\em BMC Infectious Diseases}, 21(1):70, Jan 2021.

\bibitem{rippinger2021evaluation}
Claire Rippinger, Martin Bicher, Christoph Urach, Dominik Brunmeir,
  N~Weibrecht, G~Zauner, G~Sroczynski, B~Jahn, N~M{\"u}hlberger, U~Siebert,
  et~al.
\newblock Evaluation of undetected cases during the covid-19 epidemic in
  austria.
\newblock {\em BMC Infectious Diseases}, 21(1):1--11, 2021.

\bibitem{roberts2012conceptualizing}
Mark Roberts, Louise~B Russell, A~David Paltiel, Michael Chambers, Phil McEwan,
  and Murray Krahn.
\newblock Conceptualizing a model: a report of the ispor-smdm modeling good
  research practices task force--2.
\newblock {\em Medical Decision Making}, 32(5):678--689, 2012.

\bibitem{stachowiak1973allgemeine}
Herbert Stachowiak.
\newblock {\em Allgemeine modelltheorie}.
\newblock Springer, 1973.

\bibitem{SWINERD2012118}
Chris Swinerd and Ken~R. McNaught.
\newblock Design classes for hybrid simulations involving agent-based and
  system dynamics models.
\newblock {\em Simulation Modelling Practice and Theory}, 25:118--133, 2012.

\bibitem{Wolfinger_Large_2021}
David Wolfinger, Margaretha Gansterer, Karl~F. Doerner, and Nikolas Popper.
\newblock A large neighbourhood search metaheuristic for the contagious disease
  testing problem.
\newblock {\em European Journal of Operational Research}, 304(1):169--182,
  2023.
\newblock The role of Operational Research in future epidemics/ pandemics.

\end{thebibliography}

\appendix
\section{Appendix}
\subsection{Model Family Development Time-Line}
\begin{table}[h!]
    \centering
    \begin{tabular}{p{0.7cm}|p{3.5cm}|p{3.5cm}|p{7.6cm}}
    \hline
        date & changed knowledge base & new tasks & developments\\
        \hline
     Jan 2020 & SARS-CoV-2 was detected in Europe & Decision makers required an estimate of the threat & First version of \abm was developed from a population (GEPOC\cite{bicher2018gepoc}) and influenza model\cite{miksch2011pin101}.\\
     \hdashline
     Feb 2020 & SARS-CoV-2 started spreading in Austria & Decision makers needed lockdown policy estimates & The policy module of \abm was developed.\\ 
     \hdashline
     Apr 2020 & A common agreement upon the parameters of COVID-19 and SARS-CoV-2 emerged & The need for a coordinated forecast arises among decision makers & The Austrian COVID-19 Forecasting Consortium was founded. The ensemble forecast strategy was established and A common hospital model was developed\cite{bicher2022supporting}.\\
     \hdashline
     Jul 2020  & & Policies of the first wave needed to be reevaluated & \abm was extended to cover additional policies, e.g. contact tracing\cite{Bicher_Evaluation_2021}. \textcolor{gray}{A detailed differential equation model was developed for cross-model validation of the \abm.}\\
     \hdashline
     Dec 2020  & Vaccines were announced & Vaccine prioritsation was discussed. & \abm was extended to include vaccinations and prioritisation scenarios were calculated\cite{Jahn_Targeted_2021,Bicher_Iterative_2022}. \textcolor{gray}{A vaccine supply model was developed for Austria.}\\
     \hdashline
     Jan 2021  & Variants with evolutionary advantage were detected (Alpha). & & \abm was extended for multiple variants. \textcolor{gray}{A macro model was developed to analyse the takeover of a new variant and its evolutionary advantage \cite{bicher2022time}.}\\
     \hdashline
     Jun 2021 & The first reinfections were detected. Immunity waning was confirmed. & & \abm was extended from a SIR to SIRS.\\
     \hdashline
     Jul 2021 & & Scenarios for evaluation of the current immunity level of the population of Austria were required &
     \abm was deemed too computationally expensive for a problem that didn't require its epidemiological core features. Thus, the \waningmodel was developed as a faster alternative and for cross-model validation.\\
     \hdashline
     Sep 2021 & Inhomogeneous vaccination rates affect the age-shift in epidemic waves & A forecast of the age-shift in the upcoming Delta wave was needed & The \abm could not depict the age shift in previous waves. The \agemodel was developed.\\
     \hdashline
     Nov 2021 & Hospitalisation rates dropped and feedback of hospitalised persons became negligible. & & The hospitalisation module was removed from the \abm.\\
     \hdashline
     May 2022 & & A scenario based outlook for Autumn 2022 was requested by the policymakers. & The more flexible \hospitalmodel was developed based on the original model from Gesundheit Österreich GmbH. The model was applied using forecasts from \abm and the \waningmodel.\\
    \end{tabular}
    \caption{Changing tasks and knowledge base as well as genesis of the model family. Grey parts describe vital members of the model family as well, yet we decided not to describe them in the context of this work.}  \label{tab:my_label}
\end{table}
\newpage
\subsection{\waningmodelLong - Specification}
\label{sec:spec_waningmodel}
The key concept of the model is to load and evaluate known time series of confirmed cases and vaccinations using assumptions for effectiveness and waning of immunity against infection with a certain virus strain/variant of investigation such as Delta, Omicron BA.1 or Omicron BA.2 or other factors such as severe disease progression. 

The model is developed based on the ideas of classic micro-simulation models used in health decision sciences\cite{krijkamp2018microsimulation}. The used entities in the model represent persons. They do not interact with each other and change their state via events. Table \ref{tab:parameters_waningmodel} shows a summary of the model parameters.

\paragraph{State.} The state $x=(x_1,x_2,x_3,x_{4,1},x_{4,2},\dots,x_{4,m})$ of each entity is $3+m$-dimensional: a CoV state, a detection state, a vaccination state and a series of $m$ immunity states. The immunity states decide about, whether the entity is immune against factors of interest. Necessarily, for model mechanistic purposes, these factors must include immunity against infection from all virus variants which caused cases in the regarded time period, nevertheless it may also include other observables such as severe or critical disease. The entities use the discrete state-space
\begin{align}
    x_1\in X_1&:=\{\textit{active},\textit{inactive}\}\\
    x_2\in X_2&:=\{\textit{null},\textit{detected},\textit{undetected}\}\\
    x_3\in X_3&:=\{\textit{null},\textit{1-shot},\textit{ 2-shots},\textit{3-shots}\}\\
    x_{4,i}\in X_4&:=\{\textit{false},\textit{true}\},\ \forall i\in\{1,\dots,m\}.
\end{align}
\paragraph{Initialisation.} At the start of the simulation, $N$ entities with state\\ $(\textit{false},\textit{null},\textit{null},\textit{false})$ are created - with $N$ referring to the size of the regarded population.

\paragraph{Model Input.} In order to estimate the immunisation level the model requires time series of detected cases, documented vaccinations and an estimation on the ratio of undetected cases.

\paragraph{Events.} Moreover, on the top level, we distinguish two classes of events: (a) external events, generated from external sources and (b) dynamic events generated in response on other events in the course of the simulation.

In the model, five different external events are used:
\textit{detected infection-}, \textit{undetected infection-}, \textit{first shot-}, \textit{second shot-}, and \textit{third shot event}. Generation of these events happens on a daily time base in the course of the simulation. That means, at the beginning of every new simulation day, a series of external events is generated based on model input data. These events are then also prioritised w.r. to dynamic events.
\begin{itemize}
    \item \textit{detected infection event}. These events are created daily, based on given case data per virus variant. The creation process regards a predefined delay between infection and detection of the case: Say, $n_v(t)$ is the time series for the reported positive tests for virus strain $s$ then \begin{equation}i_{d,s}(t)=\left[\sum_{i=0}^{\infty}n_s(t-i)p_{d}(i)\right]\end{equation} is the corresponding timeseries for the number of created \textit{detected infection events}. $p_{d}(i)$ is a discrete delay distribution for the time between infection and test and $[\cdot]$ indicates rounding to the nearest integer\footnote{In the actual implementation we use a stochastic rounding process to ensure, that $\sum_{t}n_s(t)\overset{P}{=}\sum_{t}i_{d,s}(t)$}.
    
    Each \textit{detected infection event}[$s$] is distributed to a random, suitable entity. An entity is regarded suitable if its CoV state is \textit{inactive} and its immunity state corresponding to infection against variant $s$ is \textit{false}.
    
    \textbf{State changes.} When the event occurs, the entity's CoV state is set to \textit{active}. Moreover, if its detection state was \textit{null} it will be set to \textit{undetected} (i.e. detected cases will keep their detected state forever).
    
    \textbf{New scheduled events.} On the known date for the reported positive test, a \textit{detection event} is scheduled. Moreover, a \textit{recovery event} parametrized with variant $s$ is scheduled given a discrete recovery distribution $p_{rd}$ for detected cases. Note, that values for $p_{rd}$ and $p_d$ must be chosen s.t. the \textit{recovery event} is always later than the \textit{detection event}.
    
    \item \textit{undetected infection event}[$s$]. These events are created and distributed daily, analogous to the creation of the \textit{detected infection event} based on given case data. Yet, a scalar detection rate parameter $0<\xi<1$ is additionally regarded. Say, $n_s(t)$ is the time series for the reported positive tests with variant $v$ then \begin{equation}i_{u,s}(t)=\left[\sum_{i=0}^{\infty}n_s(t-i)p_d(i)\frac{\xi}{1+\xi}\right]\end{equation} is the corresponding time series for the number of created \textit{undetected infection events}.
    
    \textbf{State changes.} Analogous to the \textit{detected infection event} the CoV state will wither be set to \textit{active}. If the entity's detection state was \textit{null} it will be set to \textit{undetected}.
    
    \textbf{New scheduled events.} Given the discrete recovery distribution for undetected cases $p_{ru}$ a \textit{recovery event} is scheduled parametrized with variant $s$.
    
    \item \textit{first/second/third vaccination event}. These events are created daily based on given reported vaccination data for issued first, second and third vaccine shots ($v_1(t),v_2(t),v_3(t)$). In contrast to infection events, the number and date of the events are directly taken from the corresponding time series without any delay.
    
    The events are issued to random suitable entities. An entity is suitable for
    \begin{itemize}
        \item []a \textit{first vaccination event} if its \textit{vaccination state} is \textit{null},
        \item []a \textit{second vaccination event} if its \textit{vaccination state} is \textit{1-shot} and the time to the prior \textit{first vaccination event} is at least $\delta_{1,2}$ days.
        \item []a \textit{third vaccination event} if its \textit{vaccination state} is \textit{2-shot} and the time to the prior \textit{second vaccination event} is at least $\delta_{2,3}$ days.
    \end{itemize}
    
    \textbf{State changes.} When the event occurs, the entity's vaccination state is set to \textit{1-shot}, \text{2-shots} or \textit{3-shots}.
    
    \textbf{New scheduled events.} A \textit{vaccination effect} event parametrized by the shot number is scheduled with a deterministic delay of $\delta_1$, $\delta_2$ or $\delta_3$ days, respectively.
\end{itemize}
As a direct consequence of the defined external events, the following dynamic events are introduced:
\begin{itemize}
     \item \textit{detection event}. This event is scheduled by the \textit{detected infection event} and renders the entity detected.
    
    \textbf{State changes.} The entity's detection state is set to \textit{detected}.
    
    \item \textit{recovery event}[$s$]. This event is scheduled by both \textit{infection events} and has an additional parameter $s$, the virus strain the entity recovered from. It disables the CoV state and decides about the immunity state dependent on the virus variant $s$ against all other observables.
    
    \textbf{State changes.} The entity's CoV state is set to \textit{inactive}.
    
    \textbf{New scheduled events.} 
    Dependent on the variant $s$ the entity was infected from different probabilities and waning rates are used to decide about the entity's immunity states. This is done in a two step process. Given base probabilities $b_{s,o_i},i\in\{1,\dots,m\}$, which specify the chance that recovery from variant $s$ leads to immunity against observable $o_i$, a $U(0,1)$ random number $x$ is drawn. For all $o_i$ with $o_i\geq x$, say all $o_i,i\in I$, immunity will be assigned. If the entity's immunity state against $o_i$ is not yet \textit{immune} a \textit{start immunity event} is scheduled for target $o_i$ with a delay of $0$.
    
    In a second step, waning of immunity is sampled for all $o_i,i\in I$: a $F_{s}$ distributed random number $y$ with mean $1$ is drawn. This random number is scaled via $y\cdot m_{v,o_i}$ and used as delay. Accordingly, \textit{end immunity events} are scheduled for all $o_i,i\in I$. If this scheduling process now causes two \textit{end immunity events} queued in the event list, the earlier one is cancelled.
    
    \item \textit{vaccination effect event}[shot number]. This event is scheduled by all three \textit{vaccination events}. It decides about the immunity state after the vaccine.
    
    \textbf{New scheduled events.} 
     Analogous to the recovery events, a random sampling with base probabilities $b_{v1,o_i},\dots,b_{v3,o_m}$, distribution functions $F_{v1},F_{v2},F_{v3}$, and means $m_{v1,o_1},\dots, m_{v3,o_m}$. Corresponding \textit{start-} and \textit{end immunity events} against the specific observables are scheduled.
    
    \item \textit{start immunity event}[$o$]. Scheduled either by the \textit{recovery-} or by the \textit{vaccination effect event}. Makes an entity immune against a certain observable $o$ until the next \textit{end immunity event}.
    
    \textbf{State changes.} The entity's immunity state against observable $o$ is set to \textit{true}.
    
    \item \textit{end immunity event}[$o$]. Scheduled either by the \textit{recovery-} or by the \textit{vaccination effect event}. Removes the immunity against observable $o$ from an entity.
    
    \textbf{State changes.} The entity's immunity state against observable $o$ is set to \textit{false}.
\end{itemize}
With the daily creation of the external events and the solely discrete delay distributions the model's specification as being ``event-based'' is slightly exaggerated. Yet it comes with clear advantages with respect to computation speed and output generation. The daily events give a natural rhythm to track the state of the overall simulation, which is the total number of entities sharing the same state.

\begin{table}
    \centering
    \begin{tabular}{p{2cm}|p{3.5cm}|p{5cm}}
         parameter & parameter space & usage \\\hline
         $N$ & $\mathbb{N}$ & number of entities in the model \\\hline
         $T$ & $\mathbb{N}$ & number of simulation days\\ \hline
         $s$ & known SARS-CoV-2 strains & strain, against which the immunisation level is investigated.\\ \hline
         $n(t)$, $v_1(t)$, $v_2(t)$, $v_3(t)$ & $\mathbb{N}^T$ & input time series of reported cases and first, second and third vaccinations\\\hline
         $r_s(t)$ & $[0,1]^T$ & input time series of the ratio of cases with strain $s$ among all reported cases\\\hline 
         $b_{s_j,o_i}$, $b_{v1,o_i}$, $b_{v2,o_i}$, $b_{v3,o_i}$ for all variants $s_j$ and observables $o_i$ & $[0,1]$ & basic probability that a recovery from variant $s_j$, or a first/second/third vaccination event leads to immunity against the observable  $o_i$\\ \hline
         $m_{s_j,o_i}$, $m_{v1,o_i}$, $m_{v2,o_i}$, $m_{v3,o_i}$ for all variants $s_j$ and observables $o_i$ & $\mathbb{R}^+$ & average time (in days) to lose a gained immunity against observable $o_j$ after the corresponding immunisation event. \\\hline
          $F_{s_j}$ for all variants $s_j$, $F_{v1}$, $F_{v2}$, $F_{v3}$& $X\sim F_{s_j/v_1/v_2/v_3}:$ $\mathbb{E}(X)=1$ &  distribution function for the immunity loss time with mean $1$;\\\hline
         
         $\xi$ & $(0,1)$ & detection probability of a case\\\hline
         $\delta_{1,2}$, $\delta_{2,3}$ & $\{0,1,\dots,T\}$ & minimum time between first and second/second and third shot\\\hline
         $\delta_{1}$, $\delta_{2}$, $\delta_3$ & $\{0,1,\dots,T\}$ & time between vaccination and immunisation after first, second or third shot\\\hline
         
         $p_d$ & $[0,1]^T$, $\sum_{t=1}^Tp_d=1$& discrete duration distribution between infection and reported positive test \\\hline
         $p_{rd}$ & $[0,1]^T$, $\sum_{t=1}^Tp_{rd}=1$ & discrete duration distribution between infection and recovery for detected cases\\\hline
         $p_{ru}$ & $[0,1]^T$, $\sum_{t=1}^Tp_{ru}=1$ & discrete duration distribution between infection and recovery for undetected cases\\\hline
    \end{tabular}
    \caption{Parameter table for the \waningmodel}
    \label{tab:parameters_waningmodel}
\end{table}
\subsection{\hospitalmodelLong - Specification}
\label{sec:spec_hospmodel}
The \hospitalmodelLong (short \hospitalmodel) makes use of two input time-series on daily time basis. The first time-series, $x_i,i\in\{1,\dots,T\}$, denotes the daily new confirmed SARS-CoV-2 cases, the second, $\xi_i,i\in\{1,\dots,T\}$, represents a-prior scaling factors for the hospitalisation rate. Both series must consist of positive numbers. We will explain the role of this second time-series later.

Moreover the model makes use of a time hospitalisation rate $p_i=p\xi_i$ with $p>0$ and two positive kernel vectors $\vec{a},\vec{b}\in \mathbb{R^n}$ with $\sum_{i=1}^{n}a_i=1$ and $\sum_{i=1}^{n}b_i=1$. Role of these kernels is to model a distribution of the duration between positive test and hospitalisation ($\vec{a}$) and between hospitalisation and release ($\vec{b}$). Thus, $p_ix_i$ describes the total number of hospitalisations that will originate from cases $x_i$, vector $\vec{a}p_ix_i$ can be interpreted as the number of hospital admissions at days $i,i+1,\dots,i+n$, and vector $\left(\vec{a}p_ix_i\right)_j\vec{b}$, as the number of hospital releases on days $i+j,i+j+1,\dots,i+j+n$.

With this idea we define the time-series of hospital admissions $u_i,i=1,\dots,T$, releases $v_i,i=1,\dots,T$ and occupancy $y_i,i=2,\dots,T+1$:
\begin{equation}
    u_i=\sum_{k=i-n+1}^ip\xi_kx_ka_{i-k},\quad v_i=\sum_{k=i-n+1}^iu_kb_{i-k},\quad y_{i+1}=\sum_{k=1}^{i}u_k-v_k.
\end{equation}
For convenience, we define $x_j=0 \quad \forall j\leq 0$.

Hereby the model, mapping confirmed cases onto hospital occupancy is fully defined. ICU and normal bed occupancy distinguish themselves by different rate $p$ and duration vectors. Finally, $\xi_i$ can be interpreted as an a-prior information to a dynamically changing hospitalisation rate, e.g. by a new variant with increased/decreased virulence, changing detection rate of cases, or by changing admission policies in hospitals.

\paragraph{Calibration.} Key for usage of this model is to define a proper calibration framework. Therefore, we first define that the two delay vectors $\vec{a}=\vec{a}(\mu_a),\vec{b}=\vec{b}(\mu_b)$ each depend on one parameter $\mu_a$ and $\mu_b$ which allows to scale the length of the duration. That means, increasing $\mu_a$ would lead to a higher average duration $\sum_{i=1}^{n}ia_i$, analogously for $b$. For the calibration we furthermore regard the shape of the distributions of $a$ and $b$ as fixed parameters, whereas the scales $\mu_a,\mu_b$ and the hospitalisation rate $p$ a regarded as free parameters.

In the next step we setup the time basis for the calibration. Therefore, let $ref_i,i=2,\dots,\tilde{T}$ with $\tilde{T}<T$ stand for the reference number of occupied (ICU) beds in the past. Furthermore, define $\tau \in \mathbb{N}$ so that the time basis splits in three areas:
\begin{itemize}
    \item $\{1,\dots,\tilde{T}-\tau\}$ describes a transient phase in which the output of the system is neither investigated nor used for calibration. We recommend to choose $\tilde{T}-\tau>2n$ to make sure, that the occupancy $y_{\tilde{T}-\tau+1}$ is guaranteed to be caused by the cases between times $1$ and $\tilde{T}-\tau$. This way, the choice of $y_1$ does not matter and can simply be set to $0$.
    \item $\{\tilde{T}-\tau+1,\dots,\tilde{T}\}$ is used as the calibration phase. Here,
\begin{equation}
    err=\sum_{i=\tilde{T}-\tau+1}^{\tilde{T}}\frac{(y_i-ref_i)^2}{max(1,ref_i)^2}
\end{equation}
is regarded as the error done by the simulation model.
\item $\{\tilde{T}+1,\dots,T+1\}$ is the prediction time interval. I.e. $y_i,i=\tilde{T}+1,\dots,T+1$ can be considered as the output of the model.
\end{itemize}
It remains to solve the calibration problem $\text{argmin}_{\mu_a,\mu_b,p}(err)$ to get a properly calibrated occupancy model. We typically use a Nelder-Mead simplex algorithm, since the parameter space is not particularly large, but the problem is not sufficiently smooth for a gradient-based method. See \ref{tab:parameters_hospmodel} for a summary of all model parameters.

\paragraph{Remark 1.} Since the calibration does not ensure that $y_{\tilde{T}}=ref_{\tilde{T}}$, usually the linear transformation \begin{equation}\tilde{y}_i=y_i+(ref_{\tilde{T}}-y_{\tilde{T}})\end{equation} is performed in a postprocessing step.

\paragraph{Remark 2.} The model provides a more accurate solution, if $x_i$ is limited to cases of older age groups, since they are primarily responsible for hospitalisations.

\begin{table}
    \centering
    \begin{tabular}{p{1.4cm}|p{3cm}|p{6.1cm}}
         parameter & parameter space & usage\\\hline
         $x_i$ & $\mathbb{N}^T$ & Input time-series of the confirmed cases. Has to come from surveillance systems and/or epidemic models. \\\hline
         $\xi_i$ & $(\mathbb{R}^+)^T$ & Input time-series of factors for the hospitalisation rate. If no systemic change lies within the forecasting or the calibration period, the time-series can be set to constant $1$.\\\hline
         $p$ & $\mathbb{R}^+$ with $p\xi_i<1 \forall i$ & Base hospitalisation rate. Free parameter of the calibration.
          \\\hline
         $\vec{a}(\mu_a)$ & $\mathbb{R}^+\rightarrow(\mathbb{R}^+)^n$ & Function which maps the scale parameter onto a discrete distribution for the duration from positive test to hospitalisation in days.
         \\\hline
         $\vec{b}(\mu_b)$ & $\mathbb{R}^+\rightarrow(\mathbb{R}^+)^n$ & Function which maps the scale parameter onto a discrete distribution for the duration from hospitalisation to release in days.
         \\\hline
         $\mu_a,\mu_b$ & $\mathbb{R}^+$ & Scaling parameter of the two duration distributions. Free parameters of the calibration.
         \\\hline
         $ref_i$ & $\mathbb{N}^{\tilde{T}},\tilde{T}<T$ & Reported number of occupied hospital/ICU beds in the past. Reference for the calibration. \\\hline
         $\tau$ & $\mathbb{N}$ & Number of days in the calibration window. \\\hline
    \end{tabular}
    \caption{Parameter table for the \hospitalmodel}
    \label{tab:parameters_hospmodel}
\end{table}

\subsection{\agemodelLong - Specification}
\label{sec:spec_agemodel}
In the chosen approach, population compartments like the ones in epidemiological compartment models are no longer modelled as functions of time but also of age. For this concept, we define $P(a,t)$ as the age density of a certain population compartment $P$ so that
\begin{equation}
\int_{b_1}^{b_2}P(b,t)db
\end{equation}
describes the total number of persons with age between $b_1$ and $b_2$ at time $t$ in the compartment. Clearly, the transport equation
$\frac{\partial P}{\partial t}(a,t)=\frac{\partial P}{\partial a}(a,t)$
applies to this idea with suitable boundary conditions, since the population is ageing with trivial speed.

In order to apply this idea to epidemics models, we make use of a kernel function $\kappa(a,b)$ which defines the density of the number contacts between a person with age $a$ and a person with age $b$ per time-unit. As a result,
\begin{equation}\label{eq:contacts}\lambda(P,a,t):=\int_{0}^{\infty}\kappa(a,b)P(b,t)db,\end{equation}
defines the absolute number of contacts of all persons with age $a$ with all members of compartment $P$ per time-unit at time $t$. In the SIR concept, we may use this idea to calculate all contacts between individuals in the susceptible compartment $S$ with members of the infectious compartment $I$ by the term $\frac{S(a,t)}{N}\lambda(I,a,t)$, whereas $N$ stands for the total population.

Using ths idea, we define the following integro partial differential equation (IPDE) model.

\begin{align}
\label{eq:mckendrick}
\frac{\partial S}{\partial t}(a,t)+\frac{1}{365}\frac{\partial S}{\partial a}(a,t)&=-\beta(a,t)\lambda\left(I+I_v,a,t\right)\frac{S(a,t)}{N},\\
\frac{\partial S_v}{\partial t}(a,t)+\frac{1}{365}\frac{\partial S_v}{\partial a}(a,t)&=-(1-\theta)\beta(a,t)\lambda\left(I+I_v,a,t\right)\frac{S_v(a,t)}{N},\\
\frac{\partial I}{\partial t}(a,t)+\frac{1}{365}\frac{\partial I}{\partial a}(a,t)&=\beta(a,t)\lambda\left(I+I_v,a,t\right)\frac{S(a,t)}{N}-\gamma(a)I(a,t),\\
\frac{\partial I_v}{\partial t}(a,t)+\frac{1}{365}\frac{\partial I_v}{\partial a}(a,t)&=(1-\theta)\beta(a,t)\lambda\left(I+I_v,a,t\right)\frac{S_v(a,t)}{N}-\gamma(a)I_v(a,t),\\
\frac{\partial R}{\partial t}(a,t)+\frac{1}{365}\frac{\partial R}{\partial a}(a,t)&=\gamma(a)(I(a,t)+I_v(a,t)).
\label{eq:kernel}
\end{align}
For $t=0$, the variables $S,S_v,I,I_v,R$ are defined as $L_1$-bounded, positive functions  $S_0(a)$, $S_{v,0}(a)$, $I_0(a)$, $I_{v,0}(a)$, $R_0(a)$ with compact support on $\mathbb{R}^+\cup\{0\}$.

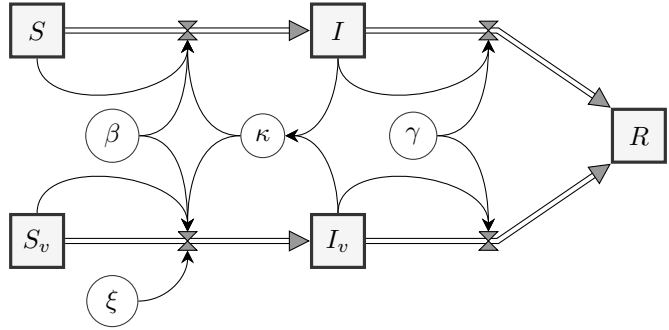
\begin{figure}
\centering
\begin{tikzpicture}[scale=2,
squarednode/.style={rectangle, draw=black!80, fill=black!4, very thick, minimum size=20},
doublearrow/.style={-{Triangle[length=8, width=8, fill=black!40]},double equal sign distance},
swingarrow/.style={-{Stealth[length=5, width=5]}},
valve/.style={}
]
\node[squarednode] (S) at (0,0) {$S$};
\node[valve] (SI) at (1,0) {};
\node[squarednode] (I) at (2,0) {$I$};
\node[circle, draw=black!80] (beta) at (0.5,-0.7) {$\beta$};
\node[circle, draw=black!80] (kappa) at (1.5,-0.7) {$\kappa$};
\node[circle, draw=black!80] (gamma) at (2.5,-0.7) {$\gamma$};
\node[squarednode] (Sv) at (0,-1.4) {$S_v$};
\node[valve] (SvIv) at (1,-1.4) {};
\node[circle, draw=black!80] (xi) at (0.5,-1.8){$\xi$};
\node[squarednode] (Iv) at (2,-1.4) {$I_v$};
\node[valve] (IR) at (3,0) {};
\node[squarednode] (R) at (4,-0.7) {$R$};
\node[valve] (IvR) at (3,-1.4) {};
\draw[doublearrow] (S.east) -- (I.west);
\draw[doublearrow] (Sv.east) -- (Iv.west);
\draw[doublearrow] (I.east) -- (IR.east) -- (R.north west);
\draw[doublearrow] (Iv.east) -- (IvR.east) --  (R.south west);
\draw[swingarrow] (beta.east) to [out=0,in=270] (SI.south);
\draw[swingarrow] (beta.east) to [out=0,in=90] (SvIv.north);
\draw[swingarrow] (xi.east) to [out=0,in=270] (SvIv.south);
\draw[swingarrow] (gamma.east) to [out=0,in=270] (IR.south);
\draw[swingarrow] (gamma.east) to [out=0,in=90] (IvR.north);

\draw[swingarrow] (S.south) to [out=270,in=270] (SI.south);
\draw[swingarrow] (Sv.north) to [out=90,in=90] (SvIv.north);
\draw[swingarrow] (I.south) to [out=270,in=0] (kappa.east);
\draw[swingarrow] (Iv.north) to [out=90,in=0] (kappa.east);
\draw[swingarrow] (kappa.west) to [out=180,in=270] (SI.south);
\draw[swingarrow] (kappa.west) to [out=180,in=90] (SvIv.north);
\draw[swingarrow] (I.south) to [out=270,in=270] (IR.south);
\draw[swingarrow] (Iv.north) to [out=90,in=90] (IvR.north);

\draw[-,fill=black!40] (SI.north west) -- (SI.north east) -- (SI.south west) -- (SI.south east) -- (SI.north west);
\draw[-,fill=black!40] (SvIv.north west) -- (SvIv.north east) -- (SvIv.south west) -- (SvIv.south east) -- (SvIv.north west);
\draw[-,fill=black!40] (IR.north west) -- (IR.north east) -- (IR.south west) -- (IR.south east) -- (IR.north west);
\draw[-,fill=black!40] (IvR.north west) -- (IvR.north east) -- (IvR.south west) -- (IvR.south east) -- (IvR.north west);
\end{tikzpicture}
\caption{Sketch of the Stock-and-Flow diagram of the IPDE model, if interpreted in the System-Dynamics-sense.}
\label{fig:flow_diagram}
\end{figure}

In this model, $S(a,t)$, $I(a,t)$ and $R(a,t)$ represent population-densities for susceptible, infectious and recovered individuals with respect to their age $a$ (in years). $S_v(a,t)$ and $I_v(a,t)$ analogously stand for vaccinated-``susceptible'' and vaccinated-infectious population densities for which the rate of infection is reduced according to the vaccine effectiveness $\theta$. Functions $\beta$ and $\gamma$ are age-dependent parameters for infection- and recovery rate of the disease, comparable with the analogous scalar parameters in the classic SIR differential equation model. Disregarding the age component, we may visualise the model via the System-Dynamics formalism\cite{forrester1970urban,forrester1987lessons} as the Stock-and-Flow diagram depicted in Figure \ref{fig:flow_diagram}. 

Moreover, the partial derivatives $\partial /\partial a$ depict the ageing process of the individuals in each cohort labelling the PDE, essentially, a transport-equation type. The factor $\frac{1}{365}$ is introduced for convenience, since age $a$ is typically measured in years whereas epidemiological rates $\beta$ and $\gamma$ are classically observed per day -- hence, $a$ is given in years, $t$ is given in days since simulation-start.

Most crucial element of the model is the introduced contact functional $\lambda((I+I_v)/N,a,t)$ which depicts the contact rate individuals with age $a$ with an infectious individual belonging to the compartments $I$ and $I_v$ per day. The overall population $N$ remains constant and can be calculated upfront via 
\begin{equation}\label{eq:N} N=\int_{0}^{\infty}S_0(b)+S_{v,0}(b)+I_0(b)+I_{v,0}(b)+R_0(b)db.\end{equation}

\begin{table}
    \centering
    \begin{tabular}{p{2cm}|p{3.5cm}|p{5cm}}
         parameter & parameter space & usage\\\hline
         $N$ & $\mathbb{N}$ & size of the investigated population \\\hline
         $\beta(a,t)$ & $\mathbb{R}^+\times \mathbb{R}^+\rightarrow \mathbb{R}^+$ & infectiousness per age and time \\\hline
         $\gamma(a)$ & $\mathbb{R}^+\rightarrow \mathbb{R}^+$ & recovery rate per age
         \\\hline
         $\kappa(a_1,a_2)$ & $\mathbb{R}^+\times \mathbb{R}^+\rightarrow \mathbb{R}^+$ & number of daily contacts of persons with age $a_1$ with persons with age $a_2$ \\\hline
         $\theta$ & $[0,1]$ & vaccine effectiveness\\\hline
         $S_0(a)$, $I_0(a)$, $S_{v,0}(a)$, $I_{v,0}(a)$, $R_{v,0}(a)$ & $C_1(\mathbb{R}^+,\mathbb{R}^+)$ with (\ref{eq:N}) & initial densities of susceptible, infectious, vaccinated, vaccinated infectious and recovered\\\hline
    \end{tabular}
    \caption{Parameter table for the \agemodel}
    \label{tab:parameters_agemodel}
\end{table}
\end{document}